\documentclass[10pt, a4paper, logo, twocolumn, copyright]{deepmind}

\pdftrailerid{redacted}

\makeatletter
\renewcommand\bibentry[1]{\nocite{#1}{\frenchspacing\@nameuse{BR@r@#1\@extra@b@citeb}}}
\makeatother

\usepackage{kantlipsum, lipsum}
\usepackage{dsfont}
\usepackage{dm-colors}
\usepackage{booktabs}
\usepackage{multirow}
\usepackage{longtable}
\usepackage{mdframed}

\usepackage[authoryear, sort&compress, round]{natbib}

\graphicspath{{figures/}}

\title{Continuous diffusion for categorical data}

\correspondingauthor{sedielem@deepmind.com}

\reportnumber{001} %

\renewcommand{\today}{} 

\newcommand{\cdcd}{CDCD}

\author[1]{Sander Dieleman}
\author[1]{Laurent Sartran}
\author[2*]{Arman Roshannai}
\author[1]{Nikolay Savinov}
\author[1]{Yaroslav Ganin}
\author[1]{Pierre H. Richemond}
\author[1]{Arnaud Doucet}
\author[3*]{Robin Strudel}
\author[1]{Chris Dyer}
\author[1]{Conor Durkan}
\author[4]{Curtis Hawthorne}
\author[1]{R\'emi Leblond}
\author[1]{Will Grathwohl}
\author[1]{Jonas Adler}

\affil[1]{DeepMind}
\affil[2]{University of Southern California}
\affil[3]{INRIA, Ecole Normale Sup\'erieure}
\affil[4]{Google Research, Brain Team}
\affil[*]{Work done while at DeepMind}

\begin{abstract}
Diffusion models have quickly become the go-to paradigm for generative modelling of perceptual signals (such as images and sound) through iterative refinement. Their success hinges on the fact that the underlying physical phenomena are continuous. For inherently discrete and categorical data such as language, various diffusion-inspired alternatives have been proposed. However, the continuous nature of diffusion models conveys many benefits, and in this work we endeavour to preserve it. We propose \cdcd, a framework for modelling categorical data with diffusion models that are continuous both in time and input space. We demonstrate its efficacy on several language modelling tasks.
\end{abstract}

\begin{document}

\maketitle

\newcommand{\expect}[2]{\mathds{E}_{{#1}} \left[ {#2} \right]}
\newcommand{\myvec}[1]{\boldsymbol{#1}}
\newcommand{\myvecsym}[1]{\boldsymbol{#1}}
\newcommand{\vx}{\myvec{x}}
\newcommand{\vy}{\myvec{y}}
\newcommand{\vz}{\myvec{z}}
\newcommand{\vtheta}{\myvecsym{\theta}}

\section{Introduction}

\begin{figure}
\begin{center}
\centerline{\includegraphics[width=0.95\columnwidth, trim=20 100 390 0, clip]{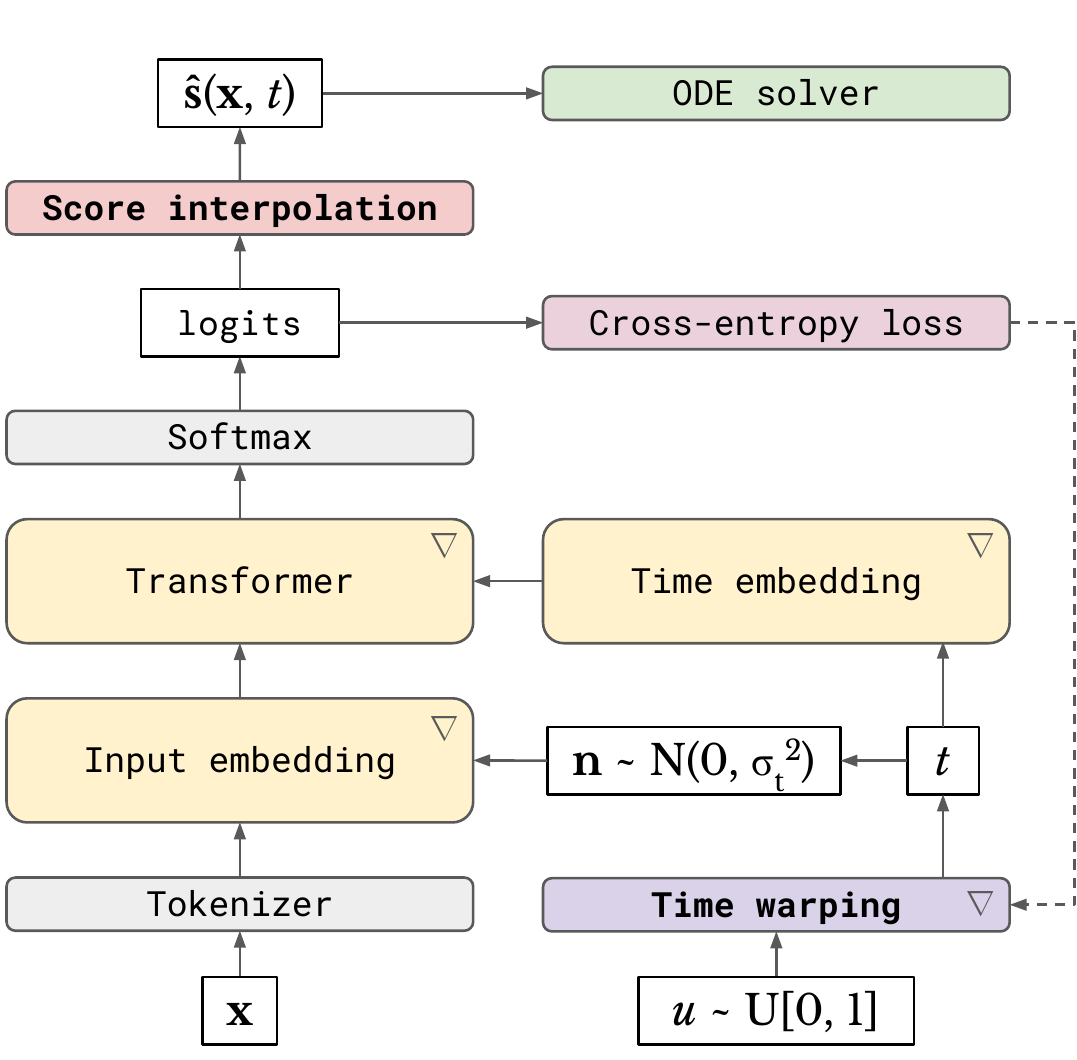}}
\caption{Overview of the \cdcd~framework. Components with learnable parameters are marked with $\nabla$. Novel components are \textbf{bolded}. The denoising model is a Transformer (without attention masking) which predicts tokens from noisy embeddings (\S\ref{sec:embeddings}) and is trained with the cross-entropy loss (\S\ref{sec:interpolation}). The input noise is time-dependent, with timesteps $t$ sampled non-uniformly during training via time warping (\S\ref{sec:time-warping}). With the predicted logits, score function estimates can be obtained through interpolation, which are used for sampling with an ODE solver. See Figure~\ref{fig:architecture-diagram} for a more detailed diagram.}
\label{fig:headline}
\end{center}
\end{figure}

Generative models have seen a rapid increase in scale and capabilities over the past few years, across many modalities, including images, audio signals, video and text~\citep{brown2020language,dhariwal2020jukebox,borsos2022audiolm,ramesh2022hierarchical,saharia2022photorealistic,ho2022imagen}.
In language modelling, the focus has been on scaling up and expanding the capabilities of autoregressive models, instigated by the development of the Transformer architecture~\citep{vaswani2017attention}.
This has resulted in general-purpose language models that are suitable for practical use.

Until recently, work on visual modalities lagged behind in terms of scale and practicability, but the development of diffusion models~\citep{sohl2015deep,song2019generative,ho2020denoising}
has resulted in a noticeable step change in capabilities. Whereas previous generative models of images were relatively inflexible and tended to produce low-resolution outputs, modern text-conditional image generators such as DALL-E 2~\citep{ramesh2022hierarchical} and Imagen~\citep{saharia2022photorealistic}
are able to produce high-resolution outputs for any conceivable textual prompt. While this trend cannot be attributed exclusively to the advent of diffusion models (models with similar capabilities that are not based on diffusion do exist, e.g. Parti~\citep{yu2022scaling}),
this new paradigm for generative modelling through iterative refinement has indisputably played a key role in the `mainstreaming' of generative models of images.

Diffusion-based language models have seen relatively little success so far. This is in part due to the discrete categorical nature of textual representations of language, which standard diffusion models are ill-equipped to deal with. As a result, several \emph{diffusion-inspired} approaches to language modelling have recently been proposed~\citep{austin2021structured,hoogeboom2021argmax,hoogeboom2021autoregressive,savinov2021step,anonymous2023diffuser}, but these depart from the diffusion modelling framework used for perceptual data in several important ways (with a few exceptions, e.g. \citet{li2022diffusion,sed}). This usually implies having to give up some of the unique capabilities of this model class, such as the ability to use classifier-free guidance to enhance conditional generation~\citep{ho2022classifier}, which has been instrumental to the success of diffusion-based text-conditional image generators.

In this paper, we study the suitability of continuous diffusion as a generative modelling paradigm for discrete categorical data, and for textual representations of language in particular. We develop a framework, \emph{Continuous diffusion for categorical data} (\cdcd), based on the diffusion framework proposed by \citet{karras2022elucidating}, which enables efficient and straightforward training of diffusion-based language models that are continuous in both time and input space, by embedding discrete tokens in Euclidean space.

Our approach very closely mirrors the training procedure for masked language models such as BERT~\citep[which are non-autoregressive and non-generative\footnote{Although several works have explored generative approaches based on masked language models~\citep{wang2019bert,ghazvininejad2019mask,goyal2021exposing,shih2022training}, they were originally introduced and are still mainly used for representation learning.}]{devlin2018bert}, and hence should appear familiar to language modelling practitioners. We hope that this will help lower the barrier to entry, and encourage researchers to explore continuous diffusion models for other domains for which categorical representations are best suited.

Our contributions are as follows:
\begin{itemize}
    \item We propose \textbf{score interpolation} as an alternative to \emph{score matching} for diffusion model training. This allows us to use the familiar cross-entropy loss function for training, which in turn enables end-to-end training of the diffusion model and the Euclidean embeddings with a single loss function;
    \item We introduce \textbf{time warping}, an active learning strategy which automatically adapts the distribution of noise levels sampled during training to maximise efficiency;
    \item We describe \cdcd, a framework for continuous diffusion models of categorical data (see Figure~\ref{fig:headline}), and explore its application to language modelling and machine translation.
\end{itemize}

\section{Diffusion models}

Diffusion models enable generative modelling via iterative denoising. Given a gradual corruption process which turns the data distribution into a simple distribution that is easy to sample from (usually an isotropic Gaussian distribution), we can train a model that learns to revert this process step by step. Each step in the reverse direction attempts to reconstruct a small amount of information that the corruption process removed. This is a much easier task than learning to generate data in a single forward pass through a model, as variational autoencoders~\citep[VAEs,][]{kingma2013auto,rezende2014stochastic} and generative adversarial networks~\citep[GANs,][]{NIPS2014_5ca3e9b1} do. Autoregressive models similarly enable decomposition of the generative modelling problem into smaller subproblems that are easier to solve. Both of these approaches to iterative refinement are compared in \S\ref{sec:diffusion-and-ar}.

\subsection{Formalism}
\label{sec:formalism}
Many different formalisms have been proposed for diffusion models, e.g. based on score matching~ \citep{song2019generative} or latent variable models~\citep{ho2020denoising}. In this work, we will follow \citet{song2020score} and use differential equations to describe the corruption process, as well as the reverse process. We believe that all these different perspectives are largely interchangeable and complementary to some degree.

\citet{song2020score} suggest modelling a diffusion process with the following stochastic differential equation (SDE):

\begin{equation}
\label{eqn:forward-sde}
\mathrm{d} \mathbf{x} = \mathbf{f}(\mathbf{x}, t) \mathrm{d}t + g(t) \mathrm{d} \mathbf{w},
\end{equation}

where $\mathbf{w}$ is the standard Wiener process, $\mathbf{f}$ is the (vector-valued) \emph{drift} coefficient, $g$ is the \emph{diffusion} coefficient and time $t$ ranges from $0$ (clean data) to $T$ (fully corrupted). The reverse process can then be described by the following SDE:

\begin{equation}
\label{eqn:reverse-sde}
\mathrm{d} \mathbf{x} = \left(\mathbf{f}(\mathbf{x}, t) - g(t)^2 \nabla_\mathbf{x} \log p_t(\mathbf{x}) \right)\mathrm{d}t + g(t) \mathrm{d} \mathbf{\bar{w}},
\end{equation}

where $\mathbf{\bar{w}}$ is the standard Wiener process in reversed time. $\mathbf{s}(\mathbf{x}, t) := \nabla_\mathbf{x} \log p_t(\mathbf{x})$ is the so-called \emph{score function}, i.e. the gradient of the density of $\mathbf{x}$ at time $t$. We can train a model to predict this quantity given $\mathbf{x}$ and $t$ using \emph{score matching}~\citep{hyvarinen2005estimation}:

\begin{equation}
\label{eqn:score-matching}
    \min \left(\hat{\mathbf{s}}(\mathbf{x}, t) - \nabla_\mathbf{x} \log p_t(\mathbf{x})\right)^2 .
\end{equation}

The estimate $\hat{\mathbf{s}}(\mathbf{x}, t)$ can then be plugged into this SDE to produce samples\footnote{We use \emph{denoising score matching} in practice~\citep{vincent2011connection}.}.

It turns out that we can instead describe the evolution of $\mathbf{x}$ over time deterministically with an ordinary differential equation (ODE):

\begin{equation}
\label{eqn:prob-flow-ode}
\mathrm{d} \mathbf{x} = \left(\mathbf{f}(\mathbf{x}, t) - \frac{1}{2} g(t)^2 \nabla_\mathbf{x} \log p_t(\mathbf{x}) \right)\mathrm{d}t.
\end{equation}

This is the \emph{probability flow ODE}, which has the same marginals $p_t(\mathbf{x})$ as the forward SDE at all timesteps $t$. This equivalence is quite powerful, because it enables us to deterministically map data examples $\mathbf{x}_0$ to latent representations $\mathbf{x}_T$, and vice versa (with $\mathbf{x}_T$ approximately following a Gaussian distribution). %

\citet{karras2022elucidating} thoroughly explored the design space of diffusion models based on the probability flow ODE formulation, and we will largely follow their recommendations here. Concretely, we will choose $\mathbf{f}(\mathbf{x}, t) = 0$ and $g(t) = \sqrt{2t}$, which yields:

\begin{equation}
\label{eqn:nvidia-ode}
\mathrm{d} \mathbf{x} = -t \nabla_\mathbf{x} \log p_t(\mathbf{x}) \mathrm{d}t.
\end{equation}

In this formulation, $t$ corresponds directly to the standard deviation of the Gaussian noise that is added to $\mathbf{x}_0$ to simulate samples from $p_t(\mathbf{x})$ (and therefore they refer to $t$ as $\sigma$ instead\footnote{See Appendix B.1 of \citet{karras2022elucidating}.}).

\subsection{Diffusion for discrete data}
When $\mathbf{x}$ is discrete, the score function is undefined. This can be worked around in two ways: we can try to define a similar iterative refinement procedure through denoising for discrete data, or we can embed $\mathbf{x}$ into a continuous space and apply continuous diffusion to the embeddings. While most of the literature has focused on the former approach (see \S\ref{sec:related-discrete-diffusion} for an overview), in this work we will explore the latter -- abandoning continuity of the input usually means that we have to forgo a lot of useful capabilities, such as classifier-free guidance, %
which we would like to keep.

Another potential advantage of lifting the discrete input into a continuous space is that it becomes possible to represent superpositions of possible outcomes at intermediate timesteps of the sampling process. In language modelling, this means that we can represent uncertainty at the individual token level: the sampling procedure only commits to specific tokens at the very end. Denoising models that operate directly in the discrete input space do not have this ability: they are only able to represent specific tokens, or the absence of a decision (through use of a `mask' token). This requirement to commit early to some subset of tokens can lead to inconsistencies in the resulting samples, which are difficult to correct retroactively\footnote{Some approaches attempt to mitigate this by allowing some proportion of tokens to be resampled multiple times~\citep{savinov2021step}.}. %

Changing the input representation to be continuous enables us to use the standard diffusion framework that has been exceptionally successful for perceptual modalities, but we should not necessarily expect it to work as well for language modelling out of the box. In fact, the most commonly used modelling setup for images implicitly reduces the loss weighting of high frequency content relative to likelihood-based models, allowing for a more efficient use of model capacity which is well aligned with human perception~\citep{song2021maximum}. Furthermore, the underlying physical phenomena that are being modelled (e.g. light intensity, air pressure) are inherently continuous. The relative ease with which diffusion models of images have been scaled to high resolution inputs can at least partially be attributed to these facts. We cannot expect to benefit from this for language modelling, as the notion of `high frequency content' is not meaningful in this setting\footnote{At least, not in the traditional sense; \citet{tamkin2020language} suggest an approach to obtain and analyse multi-scale representations of language.}. Finally, we also need to consider the impact of the choice of embedding procedure on generative modelling performance.

\subsection{Diffusion and autoregression}
\label{sec:diffusion-and-ar}

Autoregressive (AR) models currently dominate language modelling research at scale. They factorise the joint distribution over a token sequence $p(x_1, x_2, ..., x_N)$ into sequential conditionals $p(x_k|x_1, ..., x_{k-1})$ and model each of them separately (with shared parameters). This means sampling always proceeds along the direction of the sequence (i.e. from left to right, when modelling text in English), and in this case, sampling an additional token constitutes an `iterative refinement' step. Autoregression is a very natural fit for language, because it is best represented as a one-dimensional sequence of tokens. That said, the way humans tend to produce language, especially in written form, is far from linear. For many tasks, the ability to go back and refine earlier parts of the sequence, or to construct it hierarchically, is useful.

Changing the modelling paradigm to a more flexible form of iterative refinement (such as diffusion) is desirable, because the increased flexibility would facilitate new applications and could potentially reduce the computational cost of sampling. However, this is a challenging prospect because of the statistical efficiency of AR model training. Because the same parameters can be used to model all sequential conditionals, each training example provides a useful gradient signal for every step of the iterative refinement procedure. This is not the case for diffusion models, where we can only train on a single noise level for each training example. As a result, diffusion models are likely to be less data-efficient, and will converge more slowly. AR models are also able to benefit from caching of previous model activations at sampling time, which significantly reduces the computational cost of each step. Diffusion models require a full forward pass across the entire sequence at each step, which can be much more costly. %

Nonetheless, this apparent efficiency benefit of AR models does impose a rather strict constraint on the connectivity patterns within these models -- specifically, causality with respect to the input sequence. This constraint is usually implemented using some form of masking, which implies that a significant amount of computation is wasted during training. It also complicates the use of multiresolution architectures, which is very common in other domains of machine learning (e.g. computer vision). Diffusion models on the other hand are completely architecturally unrestricted, so the use of multiresolution architectures (or more exotic variants) is straightforward.

This architectural flexibility compounds with the adaptivity of the denoising procedure, which enables trading off the computational cost and sample quality at sampling time by choosing the appropriate number of iterative refinement steps, without requiring retraining or finetuning. Conversely, for AR models, the number of steps is necessarily the same as the length of the sequence to be generated\footnote{Strictly speaking, it is possible to decouple the cost of sampling from the sequence length even for autoregressive models, using probability density distillation~\citep{oord2018parallel} or alternative sampling algorithms~\citep{song2021accelerating,jayaram2021parallel}, but these approaches have not been used for language models, to the best of our knowledge.}.
More sophisticated sampling procedures for diffusion models are also being developed on a regular basis, which can be applied to existing models without any changes. %
Therefore, we believe that diffusion models for language are a worthwhile pursuit, despite their relatively reduced data efficiency.

\section{The \cdcd~framework}

We will first describe how we can train diffusion models using the familiar categorical cross-entropy loss with \emph{score interpolation}. We then show how to map categorical inputs to continuous embeddings in a way that is amenable to diffusion, which we achieve by jointly learning the embeddings and the diffusion model itself, allowing them to co-adapt. Finally, we will discuss \emph{time warping}, an active learning strategy which automatically adapts the distribution of noise levels sampled during training. Together, these components constitute a framework for continuous diffusion of categorical data, or \cdcd, which is summarised in a diagram in Figure~\ref{fig:headline}.

\subsection{Score interpolation}
\label{sec:interpolation}

Diffusion models are typically trained by minimising the \emph{score matching} objective (Equation~\ref{eqn:score-matching}), where the model learns to approximate the score function $\mathbf{s}(\mathbf{x}, t)$ in the least-squares sense. The model predictions can then be substituted directly into Equations~\ref{eqn:reverse-sde} or~\ref{eqn:prob-flow-ode} for sampling.

We observe that when the data is discrete and categorical, with tokens taken from a vocabulary of size $V$, the conditional score function $\mathbf{s}(\mathbf{x}, t | \mathbf{x}_0)$ can only assume $V$ possible values. Therefore, if we have a probabilistic prediction of $\mathbf{x}_0$, we can use it to linearly interpolate the $V$ possible values to obtain a score function estimate:

\begin{equation}
    \hat{\mathbf{s}}(\mathbf{x}, t) = \sum_{i=1}^V p(\mathbf{x}_0 = \mathbf{e}_i | \mathbf{x}, t) \mathbf{s}(\mathbf{x}, t | \mathbf{x}_0 = \mathbf{e}_i),
\end{equation}

where we have used $\mathbf{e}_i$ to represent the embedding corresponding to token $i$ in the vocabulary. Note that this is also equivalent to the expectation $\mathbb{E}_{p(\mathbf{x}_0 | \mathbf{x}, t)}\left[ \mathbf{s}(\mathbf{x}, t | \mathbf{x}_0) \right]$, which helps explain why this approach works: this expectation is also the minimiser of the score matching loss, so the global optimum is the same for score matching and score interpolation.

To obtain an estimate of $p(\mathbf{x}_0 | \mathbf{x}, t)$, we can make our model predict $V$ logits and apply a softmax nonlinearity, and minimise the categorical cross-entropy loss. This is also the standard setup used to train autoregressive language models (as well as classifiers in general), so it is well-studied and understood, and it ensures stability during training. Compared to the score matching loss, the cross-entropy loss will of course weight errors in the score function estimates differently relative to each other. This difference is important in practice, because we are only able to optimise the loss approximately (i.e. the global optimum is unlikely to be reached), and the relative weighting of the noise levels will also be different.

The conditional score function corresponding to the ODE in Equation~\ref{eqn:nvidia-ode} is given by:

\begin{equation}
    \mathbf{s}(\mathbf{x}, t | \mathbf{x}_0) = \frac{\mathbf{x}_0 - \mathbf{x}}{t^2},
\end{equation}
which is affine in $\mathbf{x}_0$~\citep{karras2022elucidating}. Therefore, the expectation $\mathbb{E}_{p(\mathbf{x}_0 | \mathbf{x}, t)}\left[ \mathbf{s}(\mathbf{x}, t | \mathbf{x}_0) \right]$ can be written as an affine function of the expectation of $\mathbf{x}_0$ itself:

\begin{equation}
   \hat{\mathbf{s}}(\mathbf{x}, t) = \frac{\mathbb{E}_{p(\mathbf{x}_0 | \mathbf{x}, t)}\left[\mathbf{x}_0\right] - \mathbf{x}}{t^2}.
\end{equation}

In other words, we can first obtain an estimate of the ground truth embedding vector $\hat{\mathbf{x}}_0 := \mathbb{E}_{p(\mathbf{x}_0 | \mathbf{x}, t)}\left[\mathbf{x}_0\right]$, and then use this to obtain a score estimate. This perspective facilitates further modifications to the score estimate, which we will discuss in \S\ref{sec:embeddings}.

\subsection{Diffusion on embeddings}
\label{sec:embeddings}

To embed the input in a continuous space, we could arbitrarily assign embeddings to different tokens, or use a representation learning technique to obtain embeddings~\citep{sed}.
However, since we are able to backpropagate gradients from the diffusion model into the embeddings using the reparameterisation trick~\citep{kingma2013auto,rezende2014stochastic}, we explore learning the embeddings and the diffusion model jointly. This yields a simpler setup, with a single shared loss function for all model parameters.

If we were to train our diffusion model with score matching, joint training would result in collapse of the embedding space. Since the model is effectively predicting the noise that is added to the embeddings, this task becomes trivial when all embeddings correspond to the same vector. This minimises the loss function, but it does not yield a useful model. Therefore, additional loss terms are necessary to prevent collapse~\citep{li2022diffusion}.

Using score interpolation, we can train the diffusion model with the cross-entropy loss instead. Since the objective is now to distinguish the true embedding from all other embeddings, given a noisy embedding as input, the model is encouraged to push the embeddings as far apart as possible (as this minimises the confounding impact of the noise). We now have the opposite problem, where joint training leads to uncontrollable growth of the embedding parameters, unless they are constrained in some way.

We could again implement such a constraint using additional loss terms, but a simpler alternative is to explicitly \emph{normalise the embedding vectors}. We find that this approach is very effective, and it yields a model that is trainable end-to-end with the cross-entropy loss, without requiring any additional terms. Concretely, we always L2-normalise the embedding vectors before they are used, but we allow the underlying parameters to vary freely. We backpropagate through the normalisation operation as needed.

We can also apply L2-normalisation to the embedding estimate $\hat{\mathbf{x}}_0$ (see \S\ref{sec:interpolation}) before calculating the score estimate, which we refer to as \emph{renormalisation}. This means that the score estimate no longer corresponds directly to the expectation $\mathbb{E}_{p(\mathbf{x}_0 | \mathbf{x}, t)}\left[ \mathbf{s}(\mathbf{x}, t | \mathbf{x}_0) \right]$, but rather a rescaled version of it\footnote{We note that L2-normalisation of the expectation resembles the calculation of the mean direction of a von Mises-Fisher distribution.}. Another possible manipulation is to \emph{clamp} $\hat{\mathbf{x}}_0$ to the nearest embedding vector in the vocabulary (in the Euclidean sense). In practice, we found both of these manipulations to work less well (see \S\ref{sec:design-decisions}) -- unlike \citet{li2022diffusion}, who found that clamping improved their results.

\subsection{Time warping}
\label{sec:time-warping}

Diffusion models are essentially denoisers that can operate at many different noise levels with a single set of shared parameters. Therefore, the degree to which the model dedicates capacity to different noise levels has a significant impact on the perceived quality of the resulting samples. We can control this by appropriately weighting the noise levels during training. The impact of different weightings has been studied extensively for diffusion models of images~\citep{nichol2021improved,song2021maximum,kingma2021variational,karras2022elucidating}.

We note that our use of the cross-entropy loss, while conveying many benefits (such as stability and end-to-end training), also changes the relative weighting of the noise levels corresponding to different timesteps $t$. Because the effects of this change are difficult to quantify, we seek to determine a \emph{time reweighting} strategy that maximises sample quality. In practice, this is best implemented through \emph{importance sampling}: rather than explicitly multiplying loss contributions from different noise levels with a weighting function $\lambda(t)$, we will instead sample $t$ from a non-uniform distribution, whose density directly corresponds to the desired relative weighting. This way, we avoid introducing significant variance in the loss estimates for timesteps $t$ for which $\lambda(t)$ is particularly large.

To sample $t$ non-uniformly in practice, we can use \emph{inverse transform sampling}: we first sample uniform samples $u \in [0, 1]$ and then \emph{warp} them using the inverse cumulative distribution function (CDF) of the distribution which corresponds to the desired weighting: $t = F^{-1}(u)$. This \emph{time warping} procedure is equivalent to time reweighting in expectation, but more statistically efficient.

To estimate the CDF $F(t)$ in question, we propose to use the following heuristic:

\begin{mdframed}[nobreak=true,align=center,linecolor=dmgray300,backgroundcolor=dmgray50,innertopmargin=10,innerbottommargin=10]
\textbf{The entropy of the model predictions should increase linearly as a function of $F(t)$.}
\end{mdframed}

This heuristic has an intuitive interpretation: it implies that the uncertainty of the model predictions (measured in bits or nats) increases at a constant rate as a function of `uniform time' $u$. Therefore, the capacity of the model should be evenly distributed across the information content of the input sequences. Furthermore, when sampling from the model, using equally spaced timesteps in uniform time will result in a gradual decrease of uncertainty with an approximately constant rate. This ensures that all sampling steps do an equal amount of work in terms of resolving uncertainty.

\begin{figure*}
\begin{center}
\centerline{\includegraphics[width=0.90\textwidth, clip]{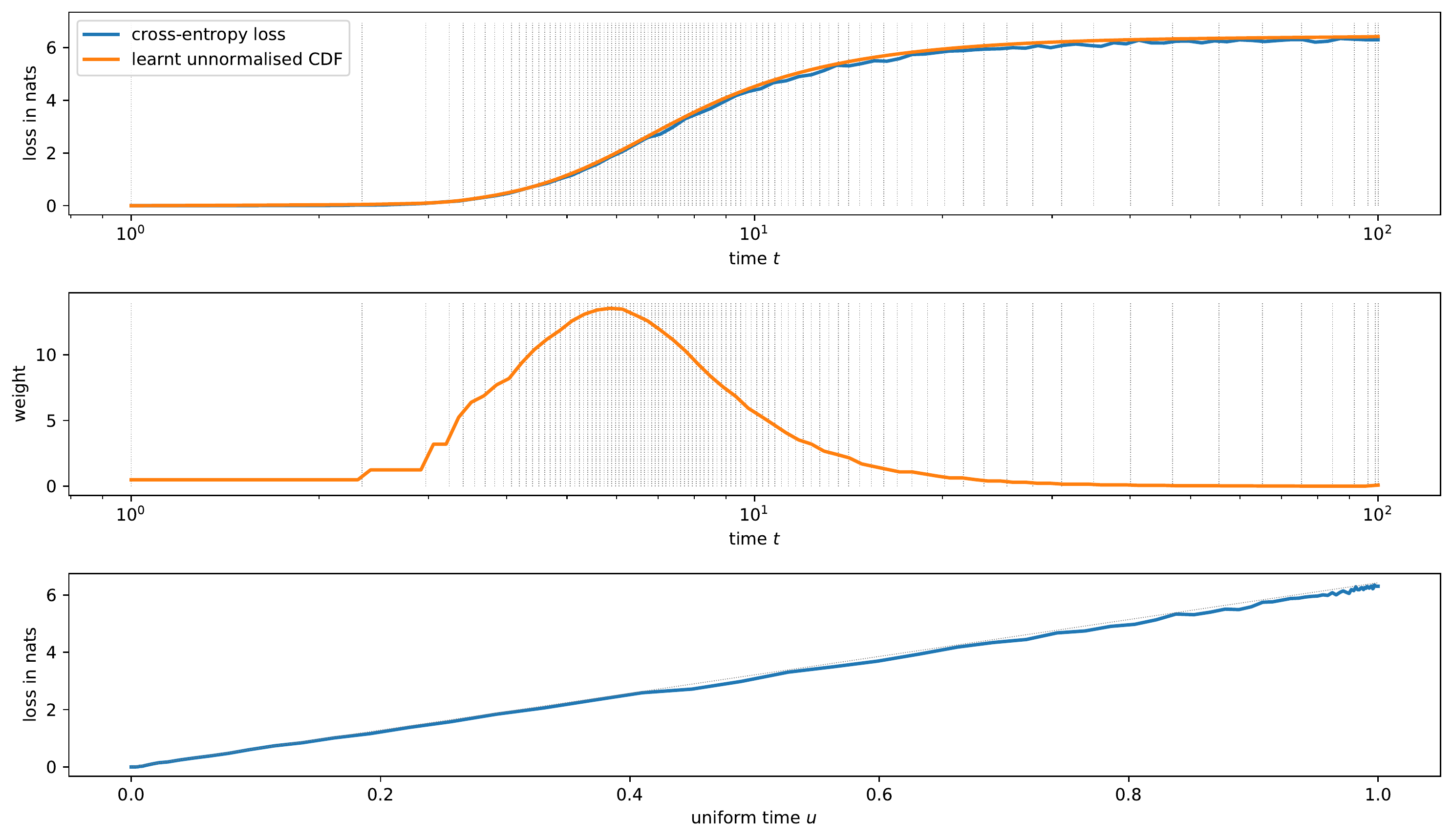}}
\caption{Time warping. Top: cross-entropy loss $\mathcal{L}(t)$ and learnt unnormalised CDF $\tilde{F}(t)$ for a fully trained model with $t_{\min}=1$ and $t_{\max}=100$. Dotted vertical lines indicate bin edges for the piecewise linear approximation (see Appendix~\ref{apx:cdf}). Middle: implied relative weighting of timesteps $\lambda(t)$ (time derivative of $F(t)$). Bottom: cross-entropy loss as a function of uniform time $u = F^{-1}(t)$.}
\label{fig:time-warping}
\end{center}
\end{figure*}

In practice, we can use the cross-entropy loss values already computed during training as a stand-in for the prediction entropy\footnote{In the limit of a model that perfectly approximates the data distribution, the prediction entropy and the cross-entropy should be the same in expectation. In practice, this is a good enough approximation even for undertrained models.}.
This means we can fit an unnormalised monotonic function $\tilde{F}(t)$ to the observed cross-entropy loss values $\mathcal{L}(t)$ (using the mean squared error loss), which we then need to normalise and invert to implement time warping:

\begin{equation}
    \min \left(\tilde{F}(t) - \mathcal{L}(t)\right)^2.
\end{equation}

This yields an \emph{active learning} strategy where noise levels are initially sampled uniformly, but as training progresses, the noise level distribution changes to focus attention towards those levels for which training is the most useful.

Inspired by \citet{muller2019neural,durkan2019neural} we parameterise $\tilde{F}(t)$ as a monotonic piecewise linear function, which is very straightforward to normalise and invert. We describe the fitting procedure in detail in Appendix~\ref{apx:cdf}. Figure~\ref{fig:time-warping} shows the cross-entropy loss as a function of $t$ for a fully trained model, along with the monotonic piecewise linear approximation $\tilde{F}(t)$ (top). It also shows the implied relative weighting of timesteps (middle) and the cross-entropy loss as a function of uniform time $u$ (bottom), which is approximately linear as a result of time warping.

Typically, we find that time warping puts most of the weight on intermediate noise levels. At very low noise levels, the denoising classification problem becomes trivial, because the embeddings corresponding to each token are easy to identify. At very high noise levels, the optimal strategy is to predict the marginal distribution of tokens (given the conditioning), which is also relatively straightforward to learn. 

\section{Diffusion language models}

\begin{figure*}
\begin{center}
\centerline{\includegraphics[width=0.58\textwidth, trim=0 0 50 310, clip]{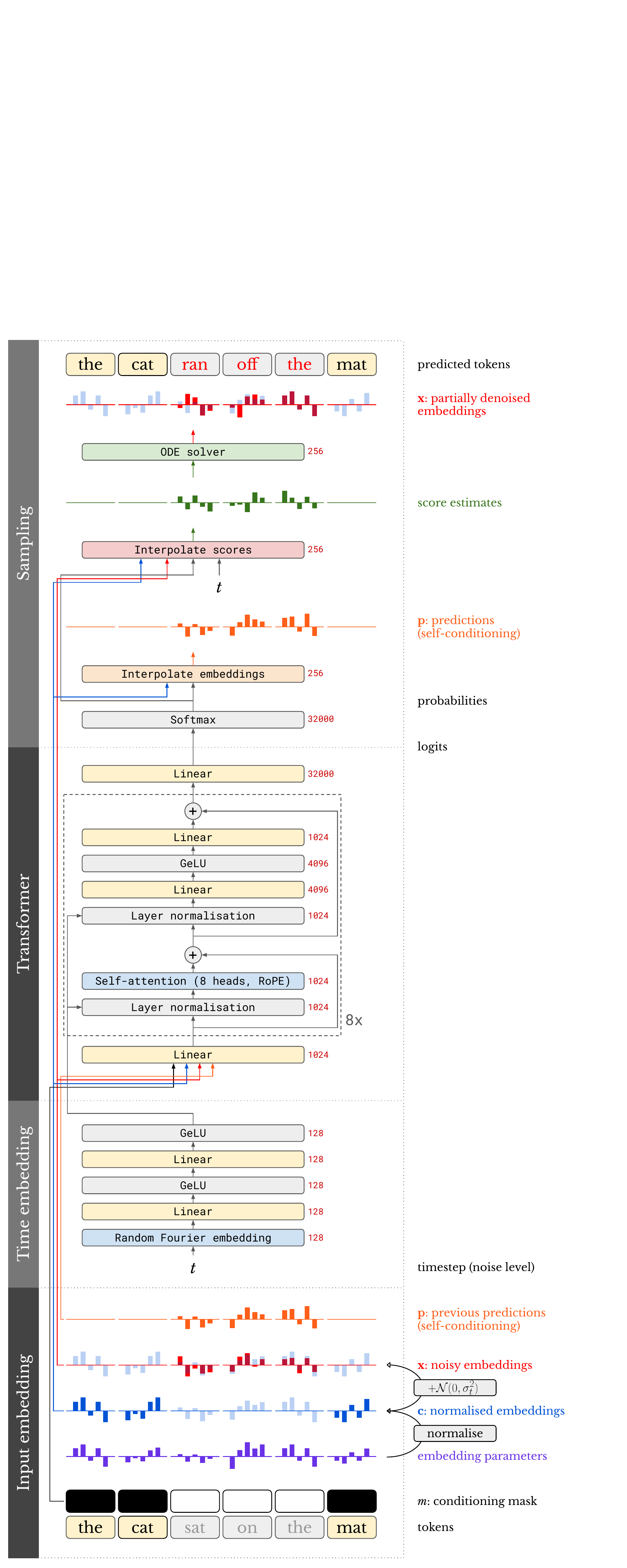}}
\caption{\small Transformer language model using the \cdcd~framework. A conditioning mask $m$ indicates which sequence positions are given (`clean') and which are to be generated (`noisy'). During training, Gaussian noise is added to the embeddings for noisy positions. The Transformer input consists of a timestep ($t$) embedding, along with the concatenation of the mask $m$, token embeddings $\mathbf{c}$ for clean positions, and noisy embeddings $\mathbf{x}$ and previous predictions $\mathbf{p}$ for noisy positions. The output logits are used to interpolate the vocabulary embeddings (to produce new predictions) and the corresponding score functions (to produce score estimates). The score estimates are used to partially denoise the noisy embeddings $\mathbf{x}$.}
\label{fig:architecture-diagram}
\end{center}
\end{figure*}

Using the \cdcd~framework described in the previous section, we can now construct language models for different tasks. We will describe a general Transformer~\citep{vaswani2017attention} architecture which can be used for prompt completion and infilling, using a boolean \emph{conditioning mask} to indicate which tokens in the sequence are to be sampled (`noisy'), and which tokens are given as conditioning (`clean'). The full model setup is visualised in a diagram in Figure~\ref{fig:architecture-diagram}. We will also describe an encoder-decoder model architecture for machine translation.

\subsection{Mask-conditional Transformer}
\label{sec:mask-conditional}

Since \cdcd~enables us to reduce the language modelling problem to a denoising classification task, without imposing any restrictions on the model architecture (such as causality, see \S\ref{sec:diffusion-and-ar}), we are able to use the Transformer architecture without any form of attention masking.

However, practical applications of language modelling require the ability to fix some subset of the tokens in a sequence, while generating the rest, conditioned on these given tokens. One way to achieve this would be to `clamp' certain token positions throughout the sampling procedure, by reinjecting these tokens at each step, corrupted by the appropriate level of noise. This method was proposed by \citet{song2020score} and referred to as `the replacement method' by \citet{ho2022video}. It seems attractive, because it would allow us to treat the model as fully unconditional during training. Nonetheless, it is not as effective as training the model to specifically support conditional sampling out of the box. Indeed, diffusion models for image inpainting are also more effective when trained specifically for that task~\citep{saharia2022palette}.

Therefore, we construct the input to the model by stacking three sequences:

\begin{itemize}
    \item $\mathbf{x}$: the embeddings corresponding to the noisy input sequence, with embeddings for conditioning tokens set to the zero vector;
    \item $\mathbf{c}$: the embeddings corresponding to the conditioning tokens, with embeddings for the tokens to be sampled set to the zero vector;
    \item $m$: the boolean conditioning mask, indicating which tokens are given (`clean', $m_i = 0$) and which are to be generated (`noisy', $m_i = 1$).
\end{itemize}

As discussed before in \S\ref{sec:interpolation}, the output of the model consists of a sequence of logit vectors, which correspond to the predicted probabilities of each token in the vocabulary occurring at each sequence position. When calculating the training loss, we zero out the positions for given tokens.

\paragraph{Prefix masking} Autoregressive language models naturally allow \emph{prefix conditioning}, where the start of a sequence is given and the model generates a completion. This is not as straightforward with diffusion models. Since prefix conditioning is a very general procedure for interacting with language models, we would like our models to support it. To achieve this, we randomly sample mask sequences $m$ during training which correspond to prefix conditioning. Such masks consist of a sequence of zeros of a certain length, followed by a sequence of ones, where the length of the prefix is sampled uniformly at random.

\paragraph{Fully random masking} Since diffusion models are able to iteratively refine all tokens in a sequence in parallel, we are not restricted to prefix conditioning. To enable conditioning on an arbitrary subset of sequence positions, we can sample masks $m$ fully randomly during training. Rather than sampling a mask value independently for each sequence position, we first sample a clean position count uniformly at random, and then randomly select a subset of that size from the sequence. This ensures that the model is able to support conditioning on any number of tokens.

\paragraph{Mixed masking} While fully random masking yields the most general model, supporting conditioning on any arbitrary subset of tokens in a sequence, prefix masking is sufficient to support the most common use cases for language models. Somewhat surprisingly, we find that training on an equal mixture of prefix masks and fully random masks actually slightly improves prefix completion performance (see \S\ref{sec:design-decisions}).

\subsection{Noise level conditioning}
\label{sec:noise-level-conditioning}
Diffusion models operate on inputs corrupted with varying levels of noise. These levels correspond directly to timesteps in the diffusion process. We provide the timestep as an additional input, which is incorporated into the model using \emph{conditional normalisation}: each layer normalisation operation in the model is followed by shifting and scaling the activations, with the shift and scale parameters depending on the timestep~\citep{perez2018film}.

\subsection{Self-conditioning}
\label{sec:self-conditioning}

\citet{chen2022analog} introduced \emph{self-conditioning}, which significantly improves the performance of diffusion models in certain contexts. They noted that the predictions produced by diffusion models are only used to determine the direction in which to update the noisy input, and are then discarded, which is wasteful. Giving the model direct access to the predictions it produced at the previous sampling step enables a more efficient use of model capacity, by allowing it to refine previous predictions, rather than constructing them from scratch at each step. This approach bears a strong resemblance to the \emph{unrolled denoising} strategy proposed by \citet{savinov2021step}.

To enable the model to make use of this additional input without requiring unrolling across multiple sampling steps during training, they propose a training procedure which only requires an additional forward pass on half of the batch at each training step. Following \citet{sed}, we use this procedure and find that it only increases training time by 10-15\% for our models in practice, while yielding significant performance gains.

To use self-conditioning with \cdcd, the input to the model now consists of four stacked sequences. In addition to $\mathbf{x}$, $\mathbf{c}$ and $m$ (see \S\ref{sec:mask-conditional}), we add $\mathbf{p}$, which is a sequence of embeddings found by interpolating the vocabulary embeddings using the token probabilities predicted in the previous sampling step (see \S\ref{sec:interpolation}; embeddings for conditioning tokens are set to the zero vector).

\subsection{Machine translation model}

For machine translation, we use an encoder-decoder architecture with two separate Transformer stacks. Since the conditioning (source) sequence and target sequence are separate, there is no need for masking. We adopt an architecture very similar to the original Transformer~\citep{vaswani2017attention}, with absolute positional embeddings (sinusoidal on the source side, learned on the target side) instead of RoPE, and ReLU as activation function.

Generating samples of the correct length is an important concern for translation. As we do not use any causal mask on the decoder side, we cannot simply disregard the contribution to the loss of the positions corresponding to padding tokens during training. Instead, we simply predict the whole sequence of tokens, including beginning-of-sentence (BOS), end-of-sentence (EOS), and padding tokens, up to a constant length. At sampling time, tokens past the first EOS are discarded.

In order to provide a strong conditioning signal to the decoder, we extend the conditioning described in \S\ref{sec:noise-level-conditioning} to additionally use the length of the source sequence.

\subsection{Comparison to BERT}
\label{sec:bert}
The model architecture and training procedure we have described so far is very similar to BERT~\citep{devlin2018bert}. Given the widespread use of that model, and its popularity among language modelling practitioners, we provide a side-by-side comparison.

Both models are sequence denoisers, but the nature of the noise differs. While BERT is trained on sequences corrupted by masking noise, which randomly removes a subset of tokens altogether, inputs to our model are corrupted by Gaussian noise which is added directly to the token embeddings. The input consists of a stack of multiple sequences (see \S\ref{sec:mask-conditional} and \S\ref{sec:self-conditioning}), rather than a single sequence where some of the tokens have been masked. Because the intensity of the noise varies according to the timesteps of the diffusion process, we also provide the timestep as an additional input, which is not required for BERT. Finally, to avoid uncontrollable growth of the embedding parameters, we force the embeddings to be normalised. Other than that, the model architectures are essentially identical during training, and the loss functions are the same.

\section{Related work}
\label{sec:related-work}

\subsection{Discrete diffusion}
\label{sec:related-discrete-diffusion}

Several diffusion-based and diffusion-inspired approaches have been proposed for non-autoregressive iterative refinement of discrete data, and especially for language in particular~\citep{ghazvininejad2019mask,chang2022maskgit,austin2021structured,hoogeboom2021autoregressive,hoogeboom2021argmax,savinov2021step,anonymous2023diffuser}. Replacing continuous diffusion with a discrete corruption process affords some flexibility, but it also requires forgoing several capabilities associated with the continuous paradigm, such as efficient sampling algorithms based on advanced ODE solvers, or classifier-free guidance.

More recently, several papers have proposed approaches to apply the continuous diffusion framework to discrete data. It is important to distinguish continuity in the input space from continuity of the time variable of the corruption process; for \cdcd, both are continuous. \citet{li2022diffusion}, \citet{sed} and \citet{ssdlm} all target language modelling as the primary application and use an embedding-based strategy in combination with discrete-time diffusion. \citet{campbell2022continuous} and \citet{sun2022score} propose continuous-time models for discrete input, though they do not explore the application to language modelling. \citet{meng2022concrete} propose concrete score matching, which can be applied to both discrete and continuous inputs. \citet{chen2022analog} use continuous-time diffusion applied to continuous relaxations of binary representations of the input. \citet{chen2022analog,li2022diffusion,sed} also suggested using the cross-entropy loss (in combination with other loss terms).

\subsection{Iterative refinement for machine translation}
There have been considerable efforts to apply non-autoregressive iterative refinement models to the task of machine translation. The first attempts by \citet{NAT_Gu_multimodality} already uncovered the issue of `multi-modality': since non-AR models usually predict all tokens in parallel and independently of each other, uncoordinated sampling decisions might lead to incoherencies like repeated tokens. Earlier advances in sequence-level distillation \citep{kim2016sequence} allowed to alleviate those issues, albeit at the cost of expensive training dataset creation. Another line of work introduced a latent transformer \citep{Latent_transformer_LT}: discrete latents are first sampled autoregressively, and then decoded non-autoregressively. 

LVM-DAE \citep{Lee_aka_Denoising_Autoencoders_in_papers_with_code} applied a non-autoregressive decoder multiple times to alleviate multi-modality. Insertion \citep{stern2019insertion} and Levenshtein \citep{gu2019levenshtein} transformers demonstrated good parallel decoding results on machine translation along with editing capabilities. The LVM-DAE line of work was later significantly improved upon by the CMLM \citep{ghazvininejad2019mask} and DisCo \citep{DisCo} methods via novel training and decoding procedures. A later work by \citet{CMLM_local_AR_aka_LAT} combined CMLM with local autoregression. Promising advances were achieved by another follow-up of CMLM called SMART~\citep{ghazvininejad2020semi}, further closing the gap between AR and non-AR methods. More recently, Imputer \citep{chan2020imputer,saharia2020non} achieved good results by optimizing alignment between source and target. 

\citet{kasai2020deep} questioned the speed advantage of non-AR models in machine translation by comparing them to a shallow-decoder AR baseline. SUNDAE \citep{savinov2021step} introduced step-unrolls and obtained excellent results both in machine translation and unconditional generation without relying on sequence-level distillation. \citet{huang2022improving} also eliminated distillation with an unroll-like technique. Aggressive decoding \citep{xia2022lossless} generated tokens in parallel while using AR models to verify generation and re-generate after the first deviation. DiffusER \citep{anonymous2023diffuser} used edit-based reconstructions and 2D beam search to almost completely close the gap between AR and non-AR models in machine translation. %

\subsection{Other related work}
\citet{kingma2021variational} suggest parameterising and optimising the noise schedule during training in a similar fashion to time warping (\S\ref{sec:time-warping}), though the objective is different: the goal is to minimise the variance of the diffusion loss, whereas our goal is to linearise the entropy of the model predictions.

\section{Experiments}
\label{sec:experiments}

We study the effect of various design choices, and compare language models based on \cdcd~with the standard autoregressive approach. We train mask-conditional models for tasks such as prompt completion and infilling, and encoder-decoder models for machine translation.

\subsection{Architecture and hyperparameters}
\label{sec:architecture-and-hyperparameters}
For the mask-conditional models, we use a SentencePiece tokenizer~\citep{kudo2018sentencepiece} with a vocabulary size of 32000. We use a standard pre-LN Transformer architecture~\citep{xiong2020layer} with 8 blocks, 1024 units and rotary positional encodings~\citep[RoPE]{su2021roformer}, trained for 1 million steps with batch size 1024 and sequence length 64. We use Random Fourier projections and an MLP with two layers of 128 units to produce timestep embeddings (see Figure~\ref{fig:architecture-diagram}). We also use self-conditioning~\citep{chen2022analog} and time warping. We use 100 bins for the piecewise linear unnormalised CDF $\tilde{F}(t)$ (see \S\ref{sec:time-warping} and Appendix~\ref{apx:cdf}).

We set the embedding dimensionality $d = 256$. We L2-normalise the embeddings and scale them by $\sqrt{d}$, so that each component has a standard deviation of 1. We choose $t_{\min} = 1.0$ and $t_{\max} = 300.0$. While these values are quite different from the ones suggested by \citet{karras2022elucidating} for image diffusion, we note that the discrete underlying nature of the input data makes it possible to predict the original tokens with 100\% accuracy even when noise with $\sigma = 1.0$ is added (recall that $\sigma = t$, see \S\ref{sec:formalism}). Therefore, it is not useful to consider lower noise levels, except when the embedding dimensionality $d$ is reduced. We scale the noisy embeddings by $\frac{1}{\sqrt{\sigma^2 + \sigma_{data}^2}} = \frac{1}{\sqrt{t^2 + 1}}$ before passing them into the model, so that the components again have a standard deviation of 1.

We drop out the conditioning by zeroing out the corresponding embeddings for 10\% of training examples, in order to be able to support sampling with classifier-free guidance~\citep{ho2022classifier}. We use low-discrepancy sampling~\citep{kingma2021variational} to sample uniform timesteps $u$ before applying time warping, to reduce the variance of the loss estimates. We use a mixed masking strategy (see \S\ref{sec:mask-conditional}): 50\% of the masks sampled during training are prefix masks (with the length of the prefix uniformly sampled), and the other 50\% are fully random masks (with the number of clean token positions again uniformly sampled).

We use the Adam optimiser~\citep{adam} with a learning rate of $10^{-4}$, $\beta_1 = 0.9$ and $\beta_2 = 0.99$. We use 200 Euler steps for sampling, and do not tune any sampling parameters (such as temperatures or guidance scales). While this is a large number of steps relative to the sequence length, we wanted to ensure that our measurements would not be negatively affected by discretisation errors introduced by the ODE solver.

\subsection{Evaluation}
\label{sec:evaluation}

We train mask-conditional models on the MassiveText dataset~\citep{rae2021scaling}, except for the larger model used in \S\ref{sec:completion-and-infilling}, which is trained on the publicly available C4 dataset~\citep{2019t5}. Both datasets contain web-crawled content, and are very diverse as a result. For machine translation, we train models on the WMT2014 German-English / English-German and WMT2020 Chinese-English datasets.

To evaluate our machine translation models, we follow the literature and use the BLEU score~\citep{papineni2002bleu}. For mask-conditional models, quantitative evaluation is more challenging. Following \citet{savinov2021step,sed}, we measure the likelihood of generated samples under a 1.3B parameter autoregressive language model (AR-NLL), as well as the unigram (per-token) entropy of the samples. As long as the entropy remains high enough, we found this negative log-likelihood to be strongly correlated with subjectively assessed model quality, and we made extensive use of this metric for hyperparameter exploration. To ensure a fair comparison, we always calculate these metrics using a fixed prefix mask with a prefix length of half the sequence length, regardless of the masking strategy used during training.

For experiments with a larger mask-conditional model, we also report the MAUVE metric~\citep{mauve}, which is specifically designed for open-ended text generation and has been shown to correlate with human judgement (see \S\ref{sec:completion-and-infilling}).

\subsection{Design decisions}
\label{sec:design-decisions}
\begin{figure*}
\begin{center}
\centerline{\includegraphics[width=1.0\textwidth, clip]{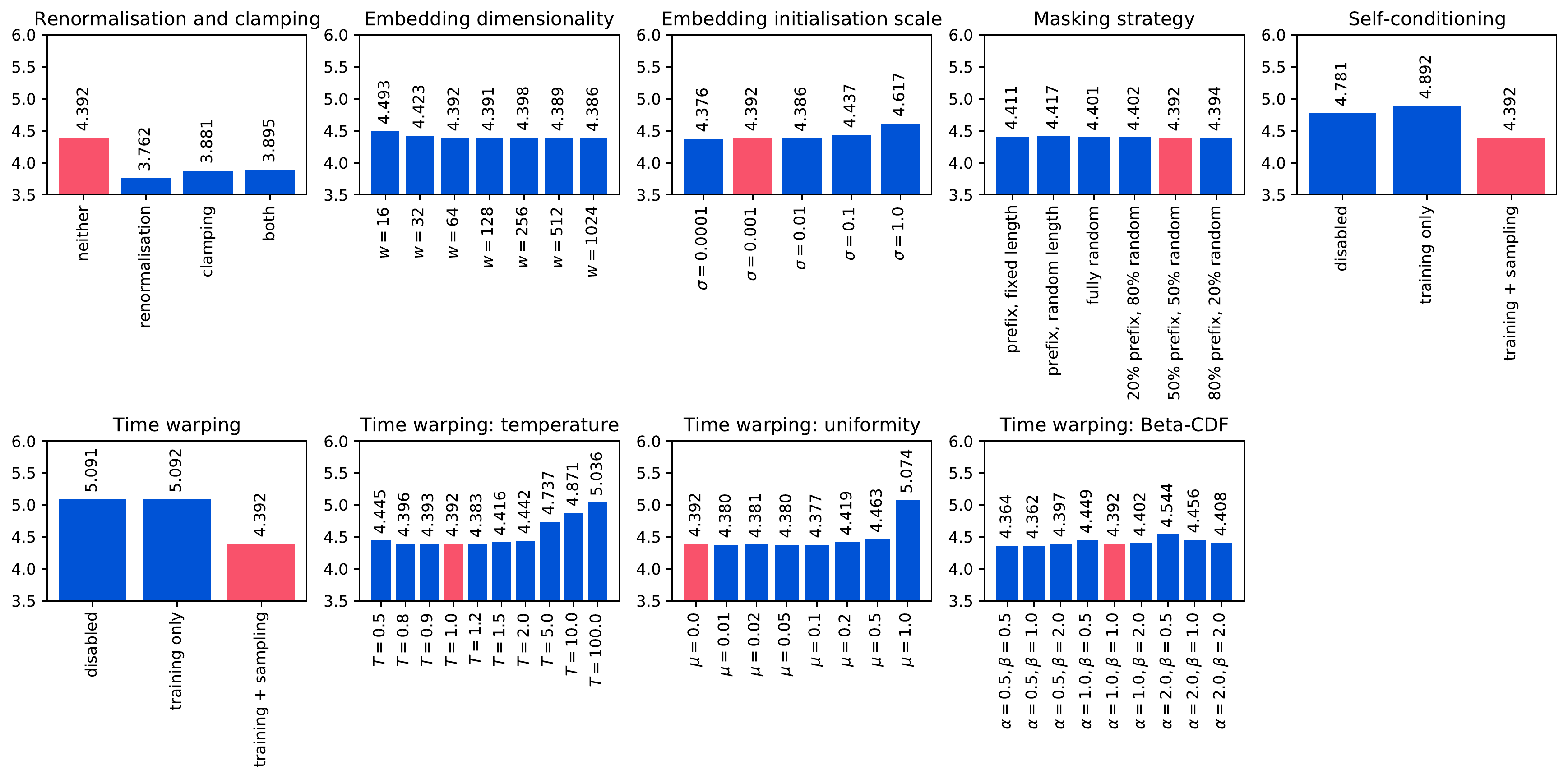}}
\caption{Design decision ablations. We report the negative log likelihood under an autoregressive language model (AR-NLL), averaged over 5 runs. See \S\ref{sec:design-decisions} and Table~\ref{tab:design-decisions} for details.}
\label{fig:design-decisions-barplots}
\end{center}
\end{figure*}

We conduct various ablation experiments to justify our architecture and hyperparameter choices, reporting both the autoregressive negative log likelihood (AR-NLL) and the unigram entropy at the token level (H) in each case. The results are summarised in Table~\ref{tab:design-decisions} and visualised in Figure~\ref{fig:design-decisions-barplots}.

\begin{table*}
    \footnotesize
    \centering
    \begin{tabular}{llcc}\toprule
    \multicolumn{2}{l}{\textbf{Ablation}} & \textbf{AR-NLL} & \textbf{Entropy} \\
    \midrule
    
    \textbf{Base model} & & $\mathbf{4.392_{\pm 0.004}}$ & $\mathbf{7.237_{\pm 0.009}}$ \\
    \midrule
    Data & & $3.389$ & $7.179$ \\
    \midrule
    
    \multirow{4}{*}{Prediction renormalisation and clamping} & \textbf{neither} & $\mathbf{4.392_{\pm 0.004}}$ & $\mathbf{7.237_{\pm 0.009}}$ \\
    & renormalisation only & $3.762_{\pm 0.003}$ & $6.946_{\pm 0.007}$ \\
    & clamping only & $3.881_{\pm 0.009}$ & $6.770_{\pm 0.020}$ \\
    & both & $3.895_{\pm 0.005}$ & $6.803_{\pm 0.011}$ \\
    \midrule
    
    \multirow{7}{*}{Embedding dimensionality ($t_{\min} = 0.1$)} & $w=16$ & $4.493_{\pm 0.006}$ & $7.268_{\pm 0.005}$ \\
    & $w=32$ & $4.423_{\pm 0.008}$ & $7.237_{\pm 0.006}$ \\
    & $w=64$ & $4.392_{\pm 0.003}$ & $7.222_{\pm 0.008}$ \\
    & $w=128$ & $4.391_{\pm 0.003}$ & $7.235_{\pm 0.007}$ \\
    & $w=256$ & $4.398_{\pm 0.009}$ & $7.248_{\pm 0.008}$ \\
    & $w=512$ & $4.389_{\pm 0.007}$ & $7.232_{\pm 0.009}$ \\
    & $w=1024$ & $4.386_{\pm 0.004}$ & $7.222_{\pm 0.003}$ \\
    \midrule
    
    \multirow{5}{*}{Embedding initialisation scale} & $\sigma=0.0001$ & $4.376_{\pm 0.004}$ & $7.226_{\pm 0.006}$ \\
    & $\mathbf{\sigma=0.001}$ & $\mathbf{4.392_{\pm 0.004}}$ & $\mathbf{7.237_{\pm 0.009}}$ \\
    & $\sigma=0.01$ & $4.386_{\pm 0.006}$ & $7.230_{\pm 0.008}$ \\
    & $\sigma=0.1$ & $4.437_{\pm 0.003}$ & $7.257_{\pm 0.003}$ \\
    & $\sigma=1.0$ & $4.617_{\pm 0.002}$ & $7.251_{\pm 0.003}$ \\
    \midrule
    
    \multirow{6}{*}{Masking strategy} & prefix, fixed length (32) & $4.411_{\pm 0.006}$ & $7.239_{\pm 0.007}$ \\
    & prefix, random length & $4.417_{\pm 0.003}$ & $7.233_{\pm 0.003}$ \\
    & fully random & $4.401_{\pm 0.011}$ & $7.206_{\pm 0.004}$ \\
    & 20\% prefix + 80\% fully random & $4.402_{\pm 0.004}$ & $7.228_{\pm 0.007}$ \\
    & \textbf{50\% prefix + 50\% fully random} & $\mathbf{4.392_{\pm 0.004}}$ & $\mathbf{7.237_{\pm 0.009}}$ \\ 
    & 80\% prefix + 20\% fully random & $4.394_{\pm 0.012}$ & $7.222_{\pm 0.009}$ \\
    \midrule
    
    \multirow{3}{*}{Self-conditioning (training and sampling)} & neither & $4.781_{\pm 0.010}$ & $7.224_{\pm 0.008}$ \\
    & training only & $4.892_{\pm 0.010}$ & $7.241_{\pm 0.008}$ \\
    & \textbf{both} & $\mathbf{4.392_{\pm 0.004}}$ & $\mathbf{7.237_{\pm 0.009}}$ \\
    \midrule
    
    \multirow{3}{*}{Time warping (training and sampling)} & neither & $5.091_{\pm 0.007}$ & $7.308_{\pm 0.006}$ \\
    & training only & $5.092_{\pm 0.013}$ & $7.311_{\pm 0.008}$ \\
    & \textbf{both} & $\mathbf{4.392_{\pm 0.004}}$ & $\mathbf{7.237_{\pm 0.009}}$ \\
    \midrule
    
    \multirow{10}{*}{Time warping: temperature} & $T=0.5$ & $4.445_{\pm 0.006}$ & $7.244_{\pm 0.003}$ \\
    & $T=0.8$ & $4.396_{\pm 0.008}$ & $7.245_{\pm 0.005}$ \\
    & $T=0.9$ & $4.393_{\pm 0.012}$ & $7.235_{\pm 0.011}$ \\
    & $\mathbf{T=1.0}$ & $\mathbf{4.392_{\pm 0.004}}$ & $\mathbf{7.237_{\pm 0.009}}$ \\
    & $T=1.2$ & $4.383_{\pm 0.005}$ & $7.226_{\pm 0.002}$ \\
    & $T=1.5$ & $4.416_{\pm 0.006}$ & $7.251_{\pm 0.007}$ \\
    & $T=2.0$ & $4.442_{\pm 0.006}$ & $7.227_{\pm 0.008}$ \\
    & $T=5.0$ & $4.737_{\pm 0.006}$ & $7.281_{\pm 0.005}$ \\
    & $T=10.0$ & $4.871_{\pm 0.009}$ & $7.275_{\pm 0.006}$ \\
    & $T=100.0$ & $5.036_{\pm 0.007}$ & $7.285_{\pm 0.006}$ \\
    \midrule
    
    \multirow{8}{*}{Time warping: uniformity} & $\mathbf{\mu=0.0}$ & $\mathbf{4.392_{\pm 0.004}}$ & $\mathbf{7.237_{\pm 0.009}}$ \\
    & $\mu=0.01$ & $4.380_{\pm 0.005}$ & $7.221_{\pm 0.009}$ \\
    & $\mu=0.02$ & $4.381_{\pm 0.006}$ & $7.222_{\pm 0.005}$ \\
    & $\mu=0.05$ & $4.380_{\pm 0.003}$ & $7.235_{\pm 0.005}$ \\
    & $\mu=0.1$ & $4.377_{\pm 0.011}$ & $7.221_{\pm 0.010}$ \\
    & $\mu=0.2$ & $4.419_{\pm 0.005}$ & $7.253_{\pm 0.005}$ \\
    & $\mu=0.5$ & $4.463_{\pm 0.010}$ & $7.249_{\pm 0.006}$ \\
    & $\mu=1.0$ & $5.074_{\pm 0.002}$ & $7.305_{\pm 0.003}$ \\
    \midrule
    
    \multirow{9}{*}{Time warping: Beta-CDF} & $\alpha=0.5, \beta=0.5$ & $4.364_{\pm 0.005}$ & $7.238_{\pm 0.005}$ \\
    & $\alpha=0.5, \beta=1.0$ & $4.362_{\pm 0.007}$ & $7.224_{\pm 0.008}$ \\
    & $\alpha=0.5, \beta=2.0$ & $4.397_{\pm 0.002}$ & $7.230_{\pm 0.006}$ \\
    & $\alpha=1.0, \beta=0.5$ & $4.449_{\pm 0.007}$ & $7.242_{\pm 0.009}$ \\
    & $\mathbf{\alpha=1.0, \beta=1.0}$ & $\mathbf{4.392_{\pm 0.004}}$ & $\mathbf{7.237_{\pm 0.009}}$ \\
    & $\alpha=1.0, \beta=2.0$ & $4.402_{\pm 0.007}$ & $7.236_{\pm 0.004}$ \\
    & $\alpha=2.0, \beta=0.5$ & $4.544_{\pm 0.011}$ & $7.236_{\pm 0.014}$ \\
    & $\alpha=2.0, \beta=1.0$ & $4.456_{\pm 0.008}$ & $7.238_{\pm 0.009}$ \\
    & $\alpha=2.0, \beta=2.0$ & $4.408_{\pm 0.009}$ & $7.225_{\pm 0.007}$ \\
    
    \bottomrule
    \end{tabular}
    \caption{Design decision ablations. We measure the negative log likelihood under an autoregressive language model (AR-NLL) and the unigram entropy (H), averaged over 5 runs, along with the standard error. Base model results are \textbf{bolded}. See \S\ref{sec:design-decisions} for details and Figure~\ref{fig:design-decisions-barplots} for a visualisation.}
    \label{tab:design-decisions}
\end{table*}

\paragraph{Renormalisation and clamping} Both manipulations of the score estimate (see \S\ref{sec:embeddings}) significantly reduce the AR-NLL, but also reduce the entropy quite a lot. Anecdotally, we also find that models using renormalisation are less amenable to improvements from sampling hyperparameter tuning, and models without renormalisation produce better samples when tuned.

Because we also make use of self-conditioning (see below), it is important to treat renormalisation and clamping as training-time hyperparameters, because they will affect the previous predictions $\mathbf{p}$ that the model receives as input (see \S\ref{sec:self-conditioning}). When not using self-conditioning, renormalisation and clamping could instead be treated as sampling hyperparameters (see \S\ref{sec:sampling}).

\paragraph{Embedding dimensionality} We reduced $t_{\min}$ from $1.0$ to $0.1$ for this experiment, because at lower embedding dimensionalities, the same amount of noise erases more information. We verified that the cross-entropy is nearly zero at $t=0.1$ for all values of $w$ we consider\footnote{The cross-entropy is significantly higher than zero at $t=1.0$ for $w=16$, for example.}. AR-NLLs initially improve with increasing dimensionality, but stabilise beyond $w=64$, thanks to time warping automatically shifting focus to the relevant noise levels (see Figure~\ref{fig:time-warping}). We used $w=256$ and $t_{\min}=1.0$ for all other experiments, including the base model, but the result obtained with $t_{\min}=0.1$ is very similar. 

\paragraph{Embedding initialisation scale} Results degrade when the scale of the initial embedding parameters is too high. We use $\sigma=0.001$ throughout to avoid this pitfall. Given its impact on performance, tuning this parameter is important, which is worth noting because weight initialisation scales are usually chosen heuristically, and generally are not treated as important hyperparameters to tune. We normalise the embeddings whenever they are used (see \S\ref{sec:embeddings}), so the scale of the underlying parameters is not relevant for inference, but clearly it significantly affects optimisation.

\paragraph{Masking strategy} We find that training with fully random masks improves results, even when evaluating using a prefix mask. We suspect that learning the embeddings becomes easier when bidirectional context is available. To ensure that the model uses enough capacity for prefix completion tasks, we use a 50-50 masking strategy for all other experiments.

\paragraph{Self-conditioning} Like \citet{sed}, we find that self-conditioning~\citep{chen2022analog} has a significant positive impact on model performance, by enabling reuse of computation from preceding sampling steps. The AR-NLL improves significantly, while the entropy stays roughly at the same level.

\paragraph{Time warping} During training, focusing on the right noise levels is clearly important, but we also find that spacing the sampling steps accordingly is essential to benefit from this improvement (see \S\ref{sec:time-warping}). We experimented with several manipulations of the warping function to verify the quality of our proposed entropy linearisation heuristic. We find that changing the temperature or the uniformity of the weighting (see Appendix~\ref{apx:cdf} and Figure~\ref{fig:temperature-uniformity}) can sometimes yield improvements, but they are relatively minor.

Targeting linearity of the loss w.r.t. uniform time $u$ implies an assumption that `all bits are equal', i.e. all information is equally important. To ensure that this is sensible, we also experimented with different target shapes of the loss as a function of $u$. To target a functional shape $S(u): [0, 1] \mapsto [0, 1]$, we minimise $\left(S(F(t)) \frac{\tilde{F}(t)}{F(t)} - \mathcal{L}(t)\right)^2$ instead of $\left(\tilde{F}(t) - \mathcal{L}(t)\right)^2$ to fit the unnormalised CDF ($S$ must be differentiable). Using the CDF of the Beta distribution for $S(u)$, we can produce target shapes with different slopes. We again find only minor improvements by deviating from $\alpha=1.0, \beta=1.0$, which corresponds to the identity function (i.e. targeting a linear shape).

\subsection{Sampling}
\label{sec:sampling}

Using the base model, we investigate the impact of various sampling hyperparameters when they are varied individually. Note that there are some significant interactions between sampling hyperparameters, which are not reflected in per-parameter experiments. We will also look at the interaction between the \emph{score temperature} and the \emph{classifier-free guidance scale} as an example of this.

\begin{table*}
    \footnotesize
    \centering
    \begin{tabular}{llccllcc}\toprule
    \multicolumn{2}{l}{\textbf{Hyperparameter}} & \textbf{AR-NLL} & \textbf{Entropy} & \multicolumn{2}{l}{\textbf{Hyperparameter}} & \textbf{AR-NLL} & \textbf{Entropy}\\
    \cmidrule(lr){1-4}\cmidrule(lr){5-8}

    \textbf{Base model} & & $\mathbf{4.392_{\pm 0.004}}$ & $\mathbf{7.237_{\pm 0.009}}$ & Data & & $3.389$ & $7.179$ \\
    \cmidrule(lr){1-4}\cmidrule(lr){5-8}

    \multirow{6}{2cm}{\# steps (Euler sampler)} & $N=10$ & $5.039_{\pm 0.009}$ & $7.281_{\pm 0.010}$ & \multirow{6}{2cm}{\# steps (Heun sampler)} & $N=5$ & $4.600_{\pm 0.006}$ & $6.955_{\pm 0.009}$ \\
    & $N=20$ & $4.775_{\pm 0.009}$ & $7.270_{\pm 0.010}$ & & $N=10$ & $4.350_{\pm 0.003}$ & $7.044_{\pm 0.007}$ \\
    & $N=50$ & $4.546_{\pm 0.005}$ & $7.249_{\pm 0.008}$ & & $N=25$ & $4.433_{\pm 0.007}$ & $7.196_{\pm 0.008}$ \\
    & $N=100$ & $4.448_{\pm 0.004}$ & $7.241_{\pm 0.008}$ & & $N=50$ & $4.414_{\pm 0.005}$ & $7.227_{\pm 0.008}$ \\
    & $\mathbf{N=200}$ & $\mathbf{4.392_{\pm 0.004}}$ & $\mathbf{7.237_{\pm 0.009}}$ & & $N=100$ & $4.380_{\pm 0.004}$ & $7.231_{\pm 0.008}$ \\
    & $N=500$ & $4.353_{\pm 0.005}$ & $7.231_{\pm 0.009}$ & & $N=250$ & $4.347_{\pm 0.004}$ & $7.229_{\pm 0.008}$ \\
    \cmidrule(lr){1-4}\cmidrule(lr){5-8}
    
    \multirow{7}{2cm}{Score temperature} & $T=0.5$ & $2.830_{\pm 0.101}$ & $6.558_{\pm 0.102}$ & \multirow{7}{2cm}{Softmax temperature} & $T=0.5$ & $3.743_{\pm 0.004}$ & $6.708_{\pm 0.006}$ \\
    & $T=0.8$ & $3.379_{\pm 0.049}$ & $7.010_{\pm 0.029}$ & & $T=0.8$ & $4.041_{\pm 0.005}$ & $6.999_{\pm 0.006}$ \\
    & $T=0.9$ & $3.838_{\pm 0.010}$ & $7.028_{\pm 0.009}$ & & $T=0.9$ & $4.201_{\pm 0.004}$ & $7.112_{\pm 0.007}$ \\
    & $T=0.95$ & $4.073_{\pm 0.003}$ & $7.118_{\pm 0.008}$ & & $T=0.95$ & $4.294_{\pm 0.003}$ & $7.174_{\pm 0.007}$ \\
    & $T=0.98$ & $4.253_{\pm 0.003}$ & $7.184_{\pm 0.008}$ & & $T=0.98$ & $4.351_{\pm 0.005}$ & $7.211_{\pm 0.008}$ \\
    & $T=0.99$ & $4.324_{\pm 0.003}$ & $7.211_{\pm 0.008}$ & & $T=0.99$ & $4.372_{\pm 0.003}$ & $7.224_{\pm 0.008}$ \\
    & $\mathbf{T=1.0}$ & $\mathbf{4.392_{\pm 0.004}}$ & $\mathbf{7.237_{\pm 0.009}}$ & & $\mathbf{T=1.0}$ & $\mathbf{4.392_{\pm 0.004}}$ & $\mathbf{7.237_{\pm 0.009}}$ \\
    \cmidrule(lr){1-4}\cmidrule(lr){5-8}
    
    \multirow{7}{2cm}{Initial noise scale} & $\sigma=0.5$ & $3.852_{\pm 0.018}$ & $7.044_{\pm 0.011}$ & \multirow{7}{2cm}{Softmax nucleus} & $p=0.5$ & $3.955_{\pm 0.004}$ & $6.965_{\pm 0.005}$ \\
    & $\sigma=0.8$ & $4.134_{\pm 0.004}$ & $7.137_{\pm 0.008}$ & & $p=0.8$ & $4.217_{\pm 0.005}$ & $7.137_{\pm 0.007}$ \\
    & $\sigma=0.9$ & $4.256_{\pm 0.002}$ & $7.184_{\pm 0.008}$ & & $p=0.9$ & $4.303_{\pm 0.003}$ & $7.188_{\pm 0.007}$ \\
    & $\sigma=0.95$ & $4.325_{\pm 0.003}$ & $7.210_{\pm 0.008}$ & & $p=0.95$ & $4.348_{\pm 0.004}$ & $7.213_{\pm 0.007}$ \\
    & $\sigma=0.98$ & $4.364_{\pm 0.004}$ & $7.226_{\pm 0.008}$ & & $p=0.98$ & $4.374_{\pm 0.004}$ & $7.228_{\pm 0.008}$ \\
    & $\sigma=0.99$ & $4.379_{\pm 0.004}$ & $7.231_{\pm 0.008}$ & & $p=0.99$ & $4.382_{\pm 0.003}$ & $7.233_{\pm 0.008}$ \\
    & $\mathbf{\sigma=1.0}$ & $\mathbf{4.392_{\pm 0.004}}$ & $\mathbf{7.237_{\pm 0.009}}$ & & $p=1.0$ & $\mathbf{4.392_{\pm 0.004}}$ & $\mathbf{7.237_{\pm 0.009}}$ \\
    \cmidrule(lr){1-4}\cmidrule(lr){5-8}

    \multirow{4}{2cm}{Step spacing (no warping)} & $\rho=1.0$ & $5.068_{\pm 0.016}$ & $7.294_{\pm 0.012}$ & \multirow{4}{2cm}{Step spacing (warping)} & $\mathbf{\rho=1.0}$ & $\mathbf{4.392_{\pm 0.004}}$ & $\mathbf{7.237_{\pm 0.009}}$ \\
    & $\rho=2.0$ & $4.631_{\pm 0.005}$ & $7.256_{\pm 0.009}$ & & $\rho=2.0$ & $4.405_{\pm 0.005}$ & $7.240_{\pm 0.009}$ \\
    & $\rho=4.0$ & $4.534_{\pm 0.004}$ & $7.248_{\pm 0.008}$ & & $\rho=4.0$ & $4.424_{\pm 0.005}$ & $7.243_{\pm 0.008}$ \\
    & $\rho=8.0$ & $4.509_{\pm 0.005}$ & $7.246_{\pm 0.009}$ & & $\rho=8.0$ & $4.443_{\pm 0.005}$ & $7.247_{\pm 0.009}$ \\
    \cmidrule(lr){1-4}\cmidrule(lr){5-8}

    \multirow{10}{2cm}{Time warping: temperature} & $T=0.5$ & $4.444_{\pm 0.004}$ & $7.259_{\pm 0.008}$ & \multirow{8}{2cm}{Time warping: uniformity} & $\mathbf{\mu=0.0}$ & $\mathbf{4.392_{\pm 0.004}}$ & $\mathbf{7.237_{\pm 0.009}}$ \\
    & $T=0.8$ & $4.396_{\pm 0.005}$ & $7.239_{\pm 0.009}$ & & $\mu=0.01$ & $4.392_{\pm 0.004}$ & $7.237_{\pm 0.009}$ \\
    & $T=0.9$ & $4.394_{\pm 0.004}$ & $7.238_{\pm 0.009}$ & & $\mu=0.02$ & $4.394_{\pm 0.004}$ & $7.236_{\pm 0.008}$ \\
    & $\mathbf{T=1.0}$ & $\mathbf{4.392_{\pm 0.004}}$ & $\mathbf{7.237_{\pm 0.009}}$ & & $\mu=0.05$ & $4.394_{\pm 0.004}$ & $7.236_{\pm 0.008}$ \\
    & $T=1.2$ & $4.396_{\pm 0.004}$ & $7.236_{\pm 0.008}$ & & $\mu=0.1$ & $4.398_{\pm 0.005}$ & $7.237_{\pm 0.008}$ \\
    & $T=1.5$ & $4.410_{\pm 0.005}$ & $7.237_{\pm 0.008}$ & & $\mu=0.2$ & $4.402_{\pm 0.004}$ & $7.237_{\pm 0.009}$ \\
    & $T=2.0$ & $4.441_{\pm 0.004}$ & $7.238_{\pm 0.008}$ & & $\mu=0.5$ & $4.435_{\pm 0.005}$ & $7.238_{\pm 0.008}$ \\
    & $T=5.0$ & $4.662_{\pm 0.006}$ & $7.255_{\pm 0.009}$ & & $\mu=1.0$ & $5.070_{\pm 0.017}$ & $7.295_{\pm 0.012}$ \\
    & $T=10.0$ & $4.833_{\pm 0.009}$ & $7.269_{\pm 0.011}$ & & & & \\
    & $T=100.0$ & $5.043_{\pm 0.014}$ & $7.291_{\pm 0.011}$ & & & & \\
    \cmidrule(lr){1-4}
    
    \multirow{5}{2cm}{CF guidance scale} & $\mathbf{\gamma=1.0}$ & $\mathbf{4.392_{\pm 0.004}}$ & $\mathbf{7.237_{\pm 0.009}}$ & & & & \\
    & $\gamma=2.0$ & $4.157_{\pm 0.007}$ & $7.198_{\pm 0.007}$ \\
    & $\gamma=4.0$ & $3.927_{\pm 0.009}$ & $7.157_{\pm 0.007}$ \\
    & $\gamma=8.0$ & $3.731_{\pm 0.005}$ & $7.118_{\pm 0.005}$ \\
    & $\gamma=16.0$ & $3.580_{\pm 0.007}$ & $7.100_{\pm 0.003}$ \\
    \cmidrule(lr){1-4}
    
    \multirow{2}{2cm}{Final prediction} & disabled & $4.394_{\pm 0.004}$ & $7.237_{\pm 0.009}$ \\
     & \textbf{enabled} & $\mathbf{4.392_{\pm 0.004}}$ & $\mathbf{7.237_{\pm 0.009}}$ \\

    \bottomrule
    \end{tabular}
    \caption{Effect of sampling hyperparameters. We measure the negative log likelihood under an autoregressive language model (AR-NLL) and the unigram entropy at the token level (H), averaged over 5 runs, along with the standard error. Base model results are \textbf{bolded}. Note that there are some significant interactions between sampling hyperparameters, which are not reflected in this per-parameter experiment. See \S\ref{sec:sampling} for details and Figure~\ref{fig:sampling-plots} for a visualisation.}
    \label{tab:sampling}
\end{table*}

\begin{figure*}
\begin{center}
\centerline{\includegraphics[width=1.0\textwidth, clip]{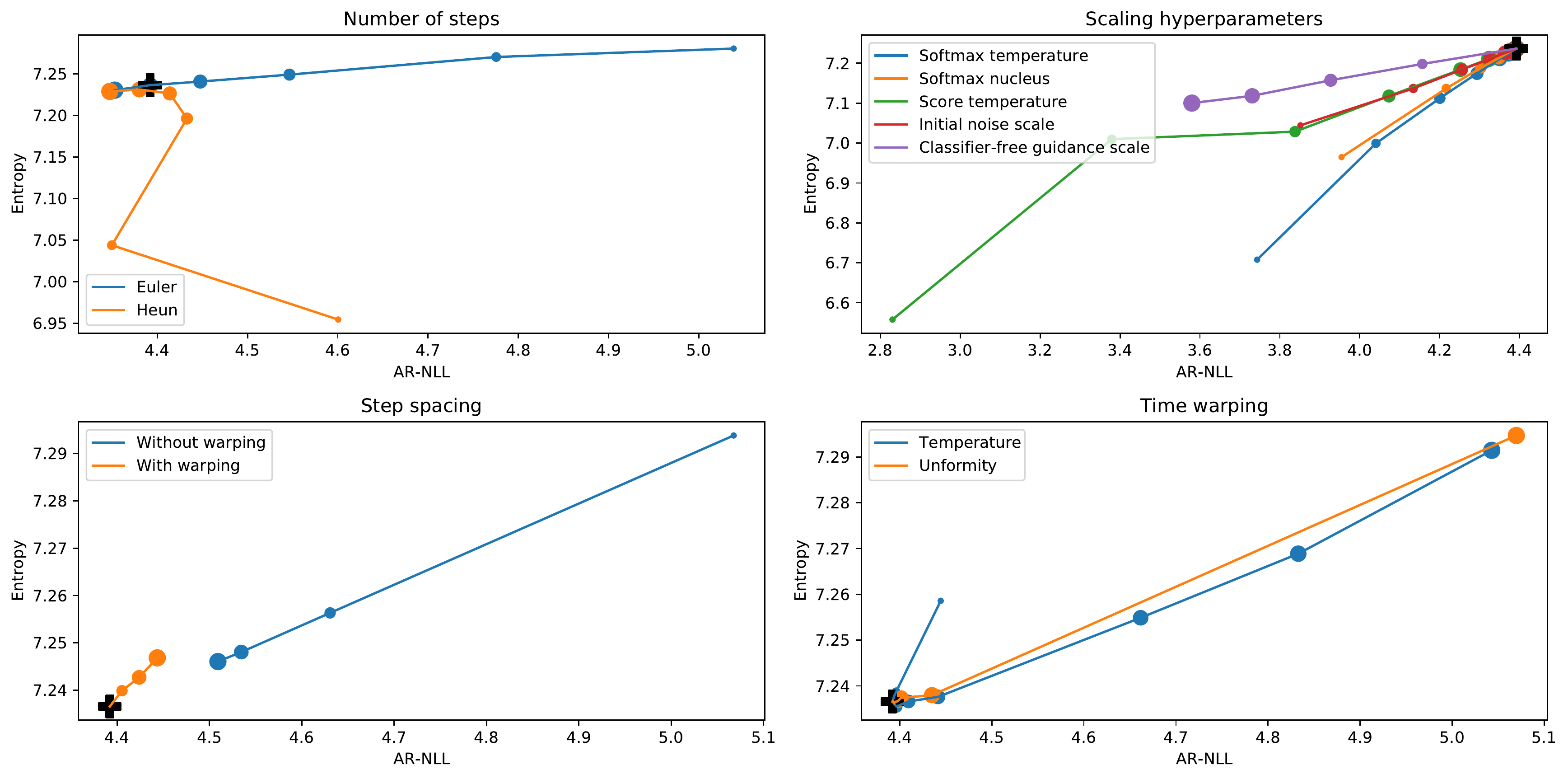}}
\caption{Effect of sampling hyperparameters. We report the negative log likelihood under an autoregressive language model (AR-NLL) and the unigram entropy at the token level (H), averaged over 5 runs. Marker size increases with hyperparameter values. The + sign corresponds to the default values. See \S\ref{sec:sampling} and Table~\ref{tab:sampling} for details.}
\label{fig:sampling-plots}
\end{center}
\end{figure*}

\paragraph{Algorithm and number of steps} We compare the Euler and Heun samplers suggested by \citet{karras2022elucidating}, reducing the number of sampling steps by a factor of two for the Heun sampler, so that the total number of required function evaluations does not change. For self-conditioning, we always use the most recent model prediction.

The Heun sampler seems to offer little benefit when using a sufficient number of steps\footnote{Note that the suitability of higher-order samplers is known to be highly dependent on other sampling hyperparameters such as the guidance scale~\citep{lu2022dpmpp}.}, and entropy degrades when the number of steps is decreased, which is not the case for the Euler sampler. We use the Euler sampler with 200 steps for most experiments, but reasonable quality can be achieved with as few as 50 function evaluations. Based on results in the image domain, it may be possible to reduce this further with stochastic samplers\footnote{Deterministic samplers have some important benefits over stochastic ones, such as support for manipulations in latent space.}, or more advanced sampling algorithms~\citep{lu2022dpm,lu2022dpmpp}.

\paragraph{Scaling hyperparameters} Many classes of generative models offer some notion of `temperature tuning'. The \cdcd~framework is particularly versatile in this regard:
\begin{itemize}
    \item scaling the score function estimate by a factor corresponds to changing the temperature of $p_t(\mathbf{x})$;
    \item the standard deviation of the initial noise can be reduced below $1.0$, which is often done for flow-based models~\citep{kingma2018glow};
    \item since the model produces a categorical probability distribution $p(\mathbf{x}_0 | \mathbf{x}, t)$, we can also manipulate this using various truncation strategies, such as temperature tuning and nucleus sampling\footnote{Nucleus \emph{sampling} is a misnomer in this case, because we never actually sample from the categorical distribution, but we can still use the same strategy to shape the logits.}~\citep{holtzman2019curious};
    \item classifier-free guidance~\citep{ho2022classifier} can be used to amplify the influence of the conditioning.
\end{itemize}

\begin{figure}
\begin{center}
\centerline{\includegraphics[width=1.0\columnwidth]{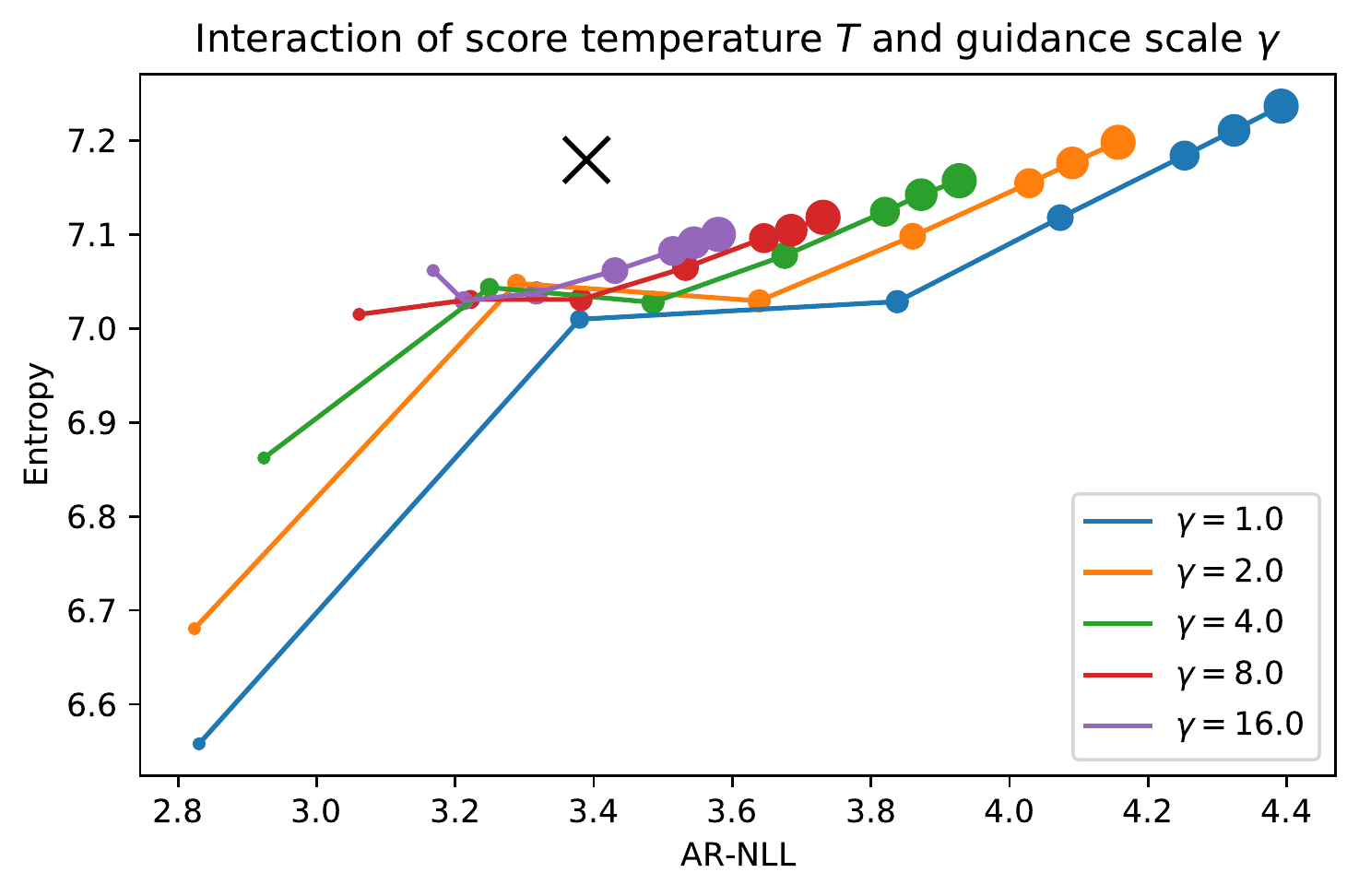}}
\caption{Interaction of the score temperature and guidance scale sampling hyperparameters. Marker size increases with score temperature. The AR-NLL and entropy of the training data are indicated with $\times$.}
\label{fig:sampling-temp-guidance}
\end{center}
\end{figure}

Out of these, we find that scaling the score temperature or changing the initial noise scale offer a strictly better trade-off than manipulating $p(\mathbf{x}_0 | \mathbf{x}, t)$. The effectiveness of changing the score temperature is surprising, because this tends to work very poorly for diffusion-based models in the visual domain. While the trade-off offered by classifier-free guidance seems even better at first glance, in practice we find that samples obtained with high guidance scales tend to contain a lot of repeated phrases. We also study the interaction between the score temperature and the guidance scale (see Figure~\ref{fig:sampling-temp-guidance}), and find that their effects are complementary to a degree.

\paragraph{Step spacing} \citet{karras2022elucidating} suggest that spacing the sampling steps non-uniformly significantly improves sample quality. For \cdcd, our use of time warping at sampling time already results in non-uniform step spacing. We compare their heuristic for $\rho = 1, 2, 4$ or $8$ with time warping, and we also investigate whether they compound, by applying time warping to non-uniformly spaced steps obtained using their heuristic.

When not using time warping at sampling time, the heuristic is clearly helpful, but it does not reach the same performance. We also find that the effects do not compound, strengthening our intuition that spacing the steps using time warping is close to optimal, because it makes the rate of decrease of uncertainty approximately constant during sampling.

\paragraph{Time warping} In \S\ref{sec:design-decisions}, we established that manipulating the warping function at training time is not particularly helpful. We can also choose to manipulate it only at sampling time, where it can still affect step spacing (but not model training). We again find that these manipulations do not have any meaningful positive effect on sample quality.

\paragraph{Final prediction} We run the model one additional time after all sampling steps are completed, and take the argmax of the predicted distributions at each sequence position to determine the sampled tokens. Instead, we could use the tokens whose embeddings are nearest to the predicted embeddings in the Euclidean sense. This works equally well, so it can save some computation, though this only becomes significant if the number of sampling steps is greatly reduced.

\subsection{Prompt completion and infilling}
\label{sec:completion-and-infilling}

We compare a 1.3B parameter model based on the \cdcd~framework with a pre-trained autoregressive model with the same architecture (24 Transformer blocks, 2048 units, 16 attention heads). Both models are trained on the C4 dataset for 600,000 steps with 524,288 tokens per batch. Due to the reduced data efficiency of diffusion model training (see \S\ref{sec:diffusion-and-ar}), it is likely that the \cdcd~model would benefit more from further training. Relative to the autoregressive model, the \cdcd~model has some extra learnable parameters in the MLP for timestep embedding, and in an initial linear layer which maps the token embeddings to the Transformer hidden state. The token embeddings themselves on the other hand account for $8\times$ fewer parameters, because we use an embedding dimensionality of $256$ instead of $2048$.

While the autoregressive model was trained with a batch size of 256 and a sequence length of 2048, we trained the \cdcd~model with a batch size of 2048 and a sequence length of 256 instead. This is partly because at this point, we are not focusing our evaluation on long-range coherence, but also because diffusion model training benefits from larger batch sizes: since noise levels are sampled on a per-sequence basis, a larger batch size yields a lower variance loss estimate. Note that the number of sampling steps is still 200, which is now smaller than the sequence length.

We take 5,000 token sequences of length 256 from the C4 validation set (rejecting shorter sequences and cropping longer ones). For each sequence, we select a random-length prefix to use as the prompt, with prompt lengths varying between 0 and 128 (half the sequence length). We sample from the autoregressive model using these prompts with nucleus sampling, for various values of $p$. We prevent the end-of-sentence (EOS) token from being sampled, to ensure that a full-length sequence is produced every time. This is necessary because the MAUVE metric is calculated on the full sequences, including the prompts, so the presence of shorter sequences would bias the results. We also sample from the \cdcd~model using the same set of prompts, with different score temperatures ($T$) and classifier-free guidance scales ($\gamma$).

\begin{figure*}
\begin{center}
\centerline{\includegraphics[width=0.96\textwidth, trim=5 0 50 5, clip]{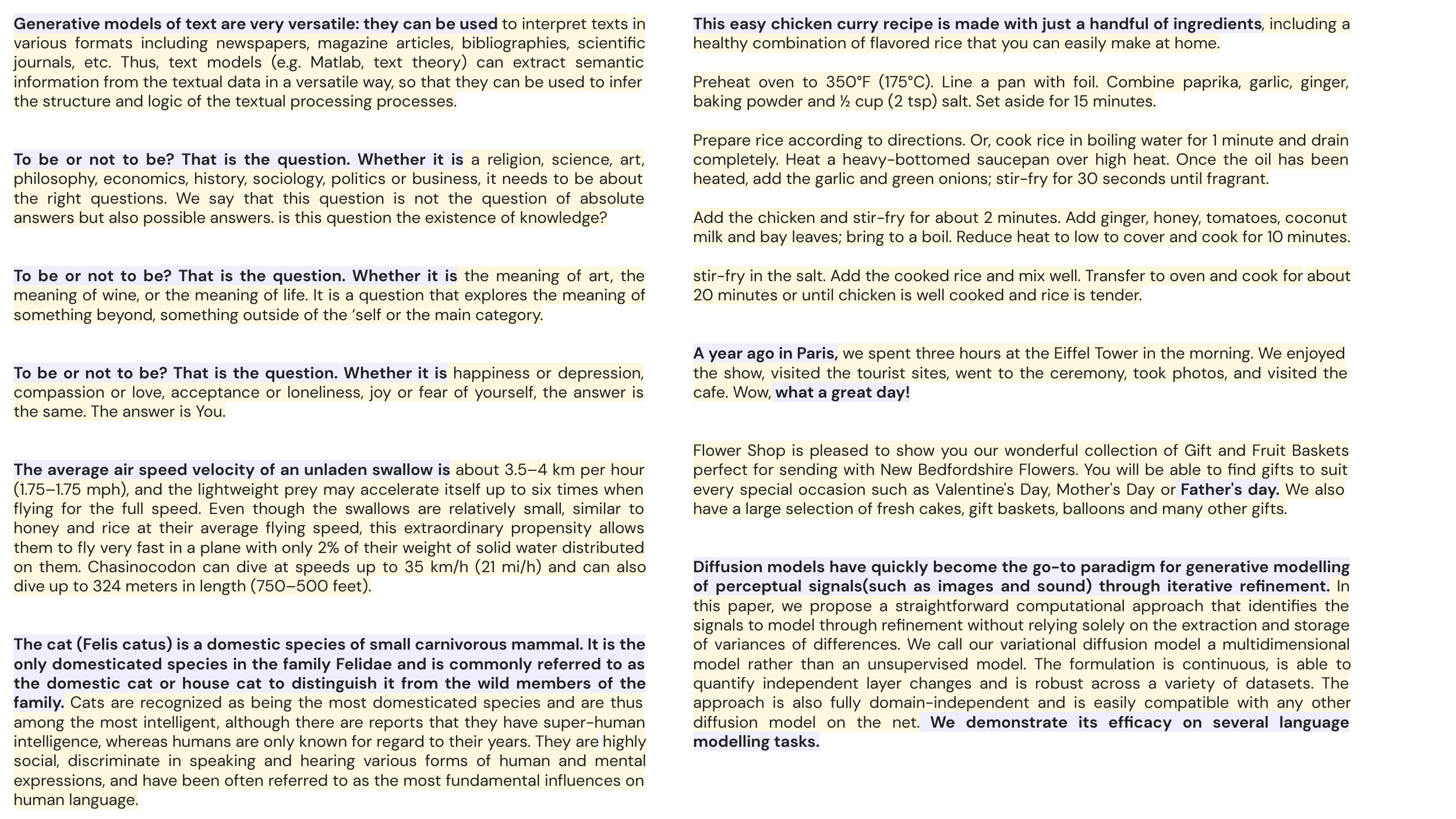}}
\caption{Completion and infilling samples from a 1.3B \cdcd~language model with score temperature $T=0.9$ and guidance scale $\gamma=1.0$ (see \S\ref{sec:completion-and-infilling}). Prompts are \textbf{bolded}. The model always produces sequences of 256 tokens, so the samples are truncated.}
\label{fig:samples}
\end{center}
\end{figure*}

The resulting completions are evaluated by comparing them with the original sequences from the dataset using MAUVE, which we report in Table~\ref{tab:prompt-completion}, alongside the AR-NLL (measured with the autoregressive model itself\footnote{These numbers are not directly comparable to those reported in Tables~\ref{tab:design-decisions} and \ref{tab:sampling}, which were obtained with a model trained on a different dataset.}) and the unigram entropy for the generated sequences. We get favourable MAUVE scores with the \cdcd~model for several settings of the score temperature $T$ and guidance scale $\gamma$. Note however that the MAUVE scores for the autoregressive samples do not seem to be convex in $p$, and it is unclear to what extent any excessive repetition introduced by increasing the guidance scale is penalised, so these numbers should be interpreted with care. Nevertheless, they provide some evidence that the \cdcd~model is able to produce compelling samples. Selected prompt completion and infilling samples are shown in Figure~\ref{fig:samples}.

\begin{table*}
    \footnotesize
    \centering
    \begin{tabular}{lllccc}\toprule
    \multicolumn{3}{l}{\textbf{Model}} & \textbf{MAUVE} & \textbf{AR-NLL} & \textbf{Entropy} \\
    \midrule
    
    Data & & & $1.000$ & $2.580$ & $7.569$ \\
    \midrule
    
    \multirow{7}{*}{Autoregressive with nucleus sampling} & $p=0.5$ & & $0.545$ & $1.469$ & $7.095$ \\
    & $p=0.8$ & & $0.900$ & $1.887$ & $7.204$ \\
    & $p=0.9$ & & $\mathbf{0.940}$ & $1.988$ & $7.235$ \\
    & $p=0.95$ & & $0.917$ & $2.016$ & $7.244$ \\
    & $p=0.98$ & & $0.928$ & $2.025$ & $7.241$ \\
    & $p=0.99$ & & $0.931$ & $2.027$ & $7.242$ \\
    & $p=1.0$ & & $0.931$ & $2.029$ & $7.242$ \\
    \midrule
    
    \multirow{30}{*}{\cdcd~with classifier-free guidance} & \multirow{4}{*}{$T=0.5$} & $\gamma=1.0$ & $0.006$ & $2.377$ & $6.591$ \\
    & & $\gamma=2.0$ & $0.006$ & $1.830$ & $6.444$ \\
    & & $\gamma=4.0$ & $0.009$ & $1.542$ & $6.532$ \\
    & & $\gamma=8.0$ & $0.013$ & $1.522$ & $6.889$ \\
    \cmidrule{2-6}
    & \multirow{4}{*}{$T=0.8$} & $\gamma=1.0$ & $0.215$ & $3.070$ & $6.726$ \\
    & & $\gamma=2.0$ & $0.523$ & $2.399$ & $7.246$ \\
    & & $\gamma=4.0$ & $0.574$ & $2.073$ & $7.488$ \\
    & & $\gamma=8.0$ & $0.555$ & $1.894$ & $7.597$ \\
    \cmidrule{2-6}
    & \multirow{4}{*}{$T=0.9$} & $\gamma=1.0$ & $0.914$ & $3.130$ & $7.353$ \\
    & & $\gamma=2.0$ & $0.939$ & $2.808$ & $7.433$ \\
    & & $\gamma=4.0$ & $0.928$ & $2.535$ & $7.510$ \\
    & & $\gamma=8.0$ & $0.884$ & $2.309$ & $7.570$ \\
    \cmidrule{2-6}
    & \multirow{4}{*}{$T=0.95$} & $\gamma=1.0$ & $0.927$ & $3.466$ & $7.491$ \\
    & & $\gamma=2.0$ & $\mathbf{0.951}$ & $3.173$ & $7.517$ \\
    & & $\gamma=4.0$ & $\mathbf{0.952}$ & $2.874$ & $7.555$ \\
    & & $\gamma=8.0$ & $0.911$ & $2.614$ & $7.594$ \\
    \cmidrule{2-6}
    & \multirow{4}{*}{$T=0.98$} & $\gamma=1.0$ & $0.907$ & $3.724$ & $7.570$ \\
    & & $\gamma=2.0$ & $0.938$ & $3.443$ & $7.576$ \\
    & & $\gamma=4.0$ & $\mathbf{0.950}$ & $3.125$ & $7.594$ \\
    & & $\gamma=8.0$ & $\mathbf{0.947}$ & $2.829$ & $7.613$ \\
    \cmidrule{2-6}
    & \multirow{4}{*}{$T=0.99$} & $\gamma=1.0$ & $0.913$ & $3.821$ & $7.595$ \\
    & & $\gamma=2.0$ & $0.925$ & $3.539$ & $7.597$ \\
    & & $\gamma=4.0$ & $\mathbf{0.953}$ & $3.217$ & $7.607$ \\
    & & $\gamma=8.0$ & $\mathbf{0.943}$ & $2.906$ & $7.621$ \\
    \cmidrule{2-6}
    & \multirow{4}{*}{$T=1.0$} & $\gamma=1.0$ & $0.889$ & $3.926$ & $7.620$ \\
    & & $\gamma=2.0$ & $0.917$ & $3.639$ & $7.618$ \\
    & & $\gamma=4.0$ & $\mathbf{0.949}$ & $3.312$ & $7.623$ \\
    & & $\gamma=8.0$ & $0.937$ & $2.986$ & $7.633$ \\
    
    \bottomrule
    \end{tabular}
    \caption{Prompt completion evaluation. We report MAUVE, AR-NLL and unigram entropy across 5,000 prefix prompts of varying length (see \S\ref{sec:completion-and-infilling}), for a 1.3B parameter autoregressive language model trained on the C4 dataset, and an equivalent \cdcd~model. MAUVE results that meet or exceed the best autoregressive result are \textbf{bolded}.}
    \label{tab:prompt-completion}
\end{table*}

\subsection{Machine translation}
\label{sec:machine-translation}

We compare a Transformer autoregressive machine translation model to a \cdcd~model with an encoder-decoder of the same size, on three translation tasks -- WMT 2014 English-German, WMT 2014 German-English, and WMT 2020 Chinese-English -- and two model sizes -- Transformer \emph{base} and \emph{big}. For each task, all models are trained on the same training set of the corresponding year, and evaluated on newstest2014 and newstest2020 accordingly. The standard validation sets are used for early stopping and for tuning the hyperparameters of \cdcd. We train two shared SentencePiece~\citep{kudo2018sentencepiece} tokenizers (one for English/German and one for Chinese/English) of size 32,768, using the unigram method~\citep{kudo-2018-subword}, with byte fallback.

For the autoregressive baseline, we use the hyperparameters described by \citet{vaswani2017attention}, except for the following: we tie the source/target/output embeddings following~\citet{press-wolf-2017-using}, we use the fixed vocabulary size indicated above, and, for Chinese-English only, we use a beam size of 6, and a length penalty $\alpha = 0.6$. The \cdcd~models are trained for 2M steps, with a batch size of 512 for base and 1024 for big, and a sequence length of 160. We use the same dropout values as for the autoregressive baseline. At sampling time, swe set the score temperature $T=0.8$ and classifier-free guidance scales $\gamma = 4.0$ for English-German and German-English, and $\gamma = 8.0$ for Chinese-English, found using a manual hyperparameter search on the validation sets.

Using a diffusion framework provides a natural way of approximating Minimum Bayes-Risk decoding~\citep{kumar-byrne-2004-minimum,https://doi.org/10.48550/arxiv.2108.04718} by sampling a number of hypotheses, each from a different initial noise vector, and selecting the one minimizing \emph{in expectation} a metric of interest, in our case BLEU.

We report BLEU scores computed on the test set for each model, year and language pair, using sacreBLEU\footnote{The signature is \texttt{nrefs:1|\-case:mixed|\-eff:no|\-tok:13a|\-smooth:exp|\-version:2.3.1} when English is the target language, and \texttt{nrefs:1|\-case:mixed|\-eff:no|\-tok:intl|\-smooth:exp|\-version:2.3.1} for German.}~\citep{post-2018-call}, in Table~\ref{tab:machine-translation}. Overall, \cdcd~models perform worse (between 3 and 7 BLEU points) than autoregressive counterparts of the same size. The difference is largest on English-German, where manual examination of the samples reveals that hypotheses in English sometimes appear, even though German output is expected. This certainly contributes to the poor score that the \cdcd~model obtains on this pair. It might be explained by the noisy character of the training data itself (English appearing on the German side). \cdcd~does comparatively better on Chinese-English, especially at larger scale, which may enable the model to make use of the larger training set. Finally, sampling-based MBR decoding improves translation quality, and monotonically so as a function of the number of samples, providing a gain of 0.7 to 1.8 BLEU points when using 100 samples compared to a single one.

Nonetheless, hypotheses produced by \cdcd~in our experiments tend to suffer from systematic defects, in particular repeated or missing tokens. We hypothesise that the model is not trained to recover from these errors, and that data augmentation, in the form of extra padding, or corruption, may prove beneficial. We leave exploring this direction to future work.

\begin{table*}
    \footnotesize
    \centering
    \begin{tabular}{llrrr}\toprule
    Size & Model & WMT 2014 EN-DE & WMT 2014 DE-EN & WMT 2020 ZH-EN
    \\\midrule
    \multirow{4}{*}{Transformer base} & Autoregressive & 26.2 & 30.2 & 25.2\\
    & \cdcd~($n=1$) & 19.3 & 24.9 & 20.7\\
    & \cdcd~($n=10$) & 19.7 & 25.4 & 21.7\\
    & \cdcd~($n=100$) & 20.0 & 26.0 & 22.2\\
    \midrule
    \multirow{4}{*}{Transformer big} & Autoregressive & 27.6 & 31.3 & 27.1\\
    & \cdcd~($n=1$) & 20.7 & 26.0 & 24.0\\
    & \cdcd~($n=10$) & 21.9 & 27.0 & 25.0\\
    & \cdcd~($n=100$) & 22.4 & 27.8 & 25.6\\
    \bottomrule
    \end{tabular}
    \caption{Machine translation evaluation. $n$ indicates the number of samples used to approximate Minimum Bayes-Risk decoding.}
    \label{tab:machine-translation}
\end{table*}

\section{Discussion}
\label{sec:discussion}

We have proposed \cdcd, a framework for diffusion of categorical data that is continuous both in time and input space, which enables training of non-autoregressive language models with a procedure reminiscent of BERT (see \S\ref{sec:bert}).

Our proposed approach has some advantages over autoregressive models, such as the ability to perform arbitrary infilling\footnote{Infilling is also possible with autoregressive models if they are specifically trained for this task~\citep{bavarian2022efficient}.} and a flexible sampling procedure which allows trading off sample quality and compute requirements. It also stands to benefit from diffusion model enhancements such as classifier-free guidance and improved sampling algorithms, and from the ability to deterministically map inputs to latents and vice versa (which we have not explored so far). Some limitations compared to autoregressive models were discussed in \S\ref{sec:diffusion-and-ar}. An important research question which we have not yet investigated, is how best to handle variable-length output. With autoregression, this is very naturally handled by introducing an end-of-sentence (EOS) token, which the model can predict to indicate that sampling should be halted. Diffusion models on the other hand require a fixed-size `canvas' which is iteratively refined during sampling, although inserting padding tokens at random during training (and collapsing them after sampling) could allow for some length variation~\citep{sed}, and it may also reduce the incidence of repeated tokens in the samples.

\paragraph{Model architecture}
We intentionally limited the degree of architectural exploration in this work, because the choice of architecture is largely orthogonal to the framework we have proposed. We focused on the Transformer as it is currently the canonical language model architecture. Nevertheless, we pointed out the absence of any architectural restrictions in this modelling paradigm (see \S\ref{sec:diffusion-and-ar}), which considerably simplifies the use of more intricate patterns, such as multi-resolution or Perceiver-based architectures~\citep{jaegle2021perceiver}.

We would also like to explore recent innovations for sampling from diffusion models (e.g. \citet{lu2022dpm,lu2022dpmpp}), which have so far been demonstrated chiefly in the image domain, and determine to what extent they improve sampling efficiency in the \cdcd~framework.

\paragraph{Application domains}
While we concentrated our empirical evaluation of the proposed \cdcd~framework on language tasks, none of its components are specific to language. We expect it to be suitable for any generative modelling problem involving structured categorical data. As an example, we note that latent diffusion models such as Stable Diffusion~\citep{rombach2022high} first use VQ-VAE~\citep{van2017neural} or VQ-GAN~\citep{esser2021taming} to learn a latent space in which the diffusion process is then applied, by mapping the categorical latents to the continuous embeddings from the vector quantisation bottleneck. While this works well, we hypothesise that fitting new embeddings jointly with the diffusion model could bring further improvements, especially in combination with time warping. Finally, our proposed time warping heuristic may also be useful beyond diffusion models of categorical data.

\section*{Acknowledgements}
We would like to thank Andy Brock, Bart Chrzaszcz, Noah Constant, Jeff Donahue, Douglas Eck, Dominik Grewe, Jordan Hoffmann, Patrick Kidger, Skanda Koppula, Lena Martens, Katie Millican, Ben Moran, Simon Osindero, Evan Shelhamer, Milo\v{s} Stanojevi\'c, Federico Vaggi, Björn Winckler, and the wider DeepMind team for their assistance and input.

We are also thankful to the creators and maintainers of the open source software used in this work, including Python~\citep{python}, NumPy~\citep{numpy}, SciPy~\citep{scipy}, JAX~\citep{jax}, TensorFlow~\citep{tensorflow}, the DeepMind JAX Ecosystem~\citep{dmjax}, Diffrax~\citep{kidger2021on} and Matplotlib~\citep{matplotlib}.

\bibliographystyle{abbrvnat}
\nobibliography*
\bibliography{refs}

\appendix

\section{Fitting the unnormalised CDF}
\label{apx:cdf}

\subsection{Parameterisation}
To implement time warping (see \S\ref{sec:time-warping}), we fit a monotonically increasing function to the expected loss at each timestep. It is essential to parameterise this function in a way that is easy to normalise and invert, so we can use it for inverse transform sampling.

We take inspiration from \citet{muller2019neural,durkan2019neural} and parameterise this function by dividing both the input and output range into bins. For convenience, we will assume that the normalised CDF maps the interval $[0, 1]$ to $[0, 1]$. In practice, timesteps $t$ range from $t_{\min}$ to $t_{\max}$ (with $t_{\max} > t_{\min} > 0$), but we can simply shift and scale them:

\begin{equation}
    t' := \frac{t - t_{\min}}{t_{\max} - t_{\min}} .
\end{equation}

The normalised CDF $u = F(t')$ is then parameterised using two sets of logits $l_n^t$ and $l_n^u$, $n = 1, \ldots, N$. Applying the softmax nonlinearity to both sets yields two partitions of the unit interval, which we use to define $N$ input and output regions. The sizes of the input and output bins are:

\begin{equation}
    w_n^t = \mathrm{softmax}_n(l_n^t), w_n^u = \mathrm{softmax}_n(l_n^u) .
\end{equation}

The left edges of the input and output bins are found as the cumulative sum of the sizes of all preceding bins:

\begin{equation}
    e_1^t = 0, e_n^t = \sum_{m=1}^{n-1} w_m^t, \forall n > 1 ,
\end{equation}

and similar for $e_n^u$. The right edges are found by adding the bin sizes to the left edges.

Within each region, the CDF can be evaluated by linearly interpolating the bin edges. The result is a piecewise linear function which is guaranteed to be monotonically increasing. We find that enforcing the minimal bin size to be nonzero helps ensure numerical stability, which we implement by adding a small constant to all bin sizes $w_n^t$ and $w_n^u$, and renormalising. By initialising both sets of logits $l_n^t$ and $l_n^u$ to a constant value, $F(t')$ initially represents the identity function, which is the CDF of the uniform distribution on the unit interval. We use the constant $-\log N$, so that the unnormalised CDF $\tilde{F}(t')$ is in fact normalised at the start of training.

\subsection{Fitting without normalisation}
We now wish to fit $u = F(t')$ so that it has the same shape as the expected loss as a function of time $t'$. However, $F(t')$ is normalised so that $u \in [0, 1]$, whereas the expected loss will vary between $0$ and $H$, the unigram entropy of the data. For an unconditional diffusion model, we can estimate said entropy remarkably accurately from a single batch of training data, simply by counting the frequencies of all tokens in the vocabulary to estimate the marginal token distribution. We could then use this entropy estimate to scale the loss values to the unit interval. Unfortunately, this does not work for conditional models, because in that case, the unigram entropy varies with the conditioning.

It turns out that we can easily define an unnormalised function $\tilde{F}(t')$ using the same parameterisation: instead of applying the softmax nonlinearity to the output logits $l_n^u$, we can apply the exponential function to find the bin sizes:

\begin{equation}
    w_n^t = \mathrm{softmax}_n(l_n^t), w_n^u = \exp(l_n^u) .
\end{equation}

The bin edges are now free to assume any positive real value, rather than being constrained to the unit interval. $\tilde{F}(t')$ can therefore be fit to the observed loss values directly. The result of this is shown in Figure~\ref{fig:time-warping} (top) for a fully trained model. To ensure that we fit the expectation of the loss at each timestep, we minimise the mean squared error (MSE) loss with respect to the per-sequence loss values observed during training. Once fit, we can obtain the normalised function $F(t') \propto \tilde{F}(t')$ by applying the softmax nonlinearity to the output logits instead of the exponential function, as before.

\subsection{Inverse and derivative}
Apart from being trivial to normalise, it is also very easy to invert the learnt CDF, simply by switching the roles of the input and output logits $l_n^t$ and $l_n^u$. We can also easily evaluate the derivative of the CDF (i.e. the probability density function), which is piecewise constant: within each bin, it is equal to the ratio of the output and input bin size, $w_n^u / w_n^t$ (Figure~\ref{fig:time-warping}, middle).

\subsection{Importance weighting}
Since we are using this fitting mechanism to change the distribution of sampled timesteps over the course of training, we have created a feedback loop, because the training data for $F(t')$ will itself become biased towards oversampled timesteps over the course of training. To compensate for this, we can use importance weights, which correspond to the reciprocal of the derivative of the CDF (or equivalently, the derivative of the inverse CDF). We take care not to backpropagate gradients through the time warping operation and the importance weights. For additional stability, we use an exponential moving average of the parameters $l_n^t$ and $l_n^u$ when performing the warping, to limit the rate of change of the training distribution, though empirically we have found that this is not strictly necessary.

\subsection{Warping sampling timesteps}
For sampling, we find that using steps that are linearly spaced in uniform time, and subsequently warped, works considerably better than using uniformly spaced timesteps. Using warping for both training (to change the distribution of noise levels on the fly) and sampling (to space the timesteps nonlinearly) yields the best results. We hypothesise that decreasing the entropy at a constant rate from step to step is also a useful heuristic for sampling.

\subsection{Temperature and uniformity}
The piecewise linear parameterisation also enables some useful distribution manipulations: we can easily change the temperature $T$ of the distribution represented by the CDF, by changing the output bin sizes:

\begin{equation}
    {w'}_n^u \propto w_n^u \cdot \left(\frac{w_n^u}{w_n^t}\right)^{T^{-1} - 1} .
\end{equation}

To see why this corresponds to a temperature change, recall that the PDF is piecewise constant, and the values it assumes (up to a normalisation constant) are given by:

\begin{equation}
    \frac{{w'}_n^u}{w_n^t} \propto \frac{w_n^u}{w_n^t} \cdot \left(\frac{w_n^u}{w_n^t}\right)^{T^{-1} - 1} = \left(\frac{w_n^u}{w_n^t}\right)^{T^{-1}}.
\end{equation}

Similarly, we can also derive the CDF of a mixture of the distribution with a uniform distribution, with mixture weight $\mu$:

\begin{equation}
    {w''}_n^u = (1 - \mu) \cdot w_n^u + \mu \cdot w_n^t .
\end{equation}

Note that when the corresponding input and output bin sizes are all equal to each other (i.e. $w_n^u = w_n^t, \forall n$), we always obtain the identity function, which corresponds to the CDF of the uniform distribution on the unit interval.

\begin{figure}
\begin{center}
\centerline{\includegraphics[width=1.0\columnwidth]{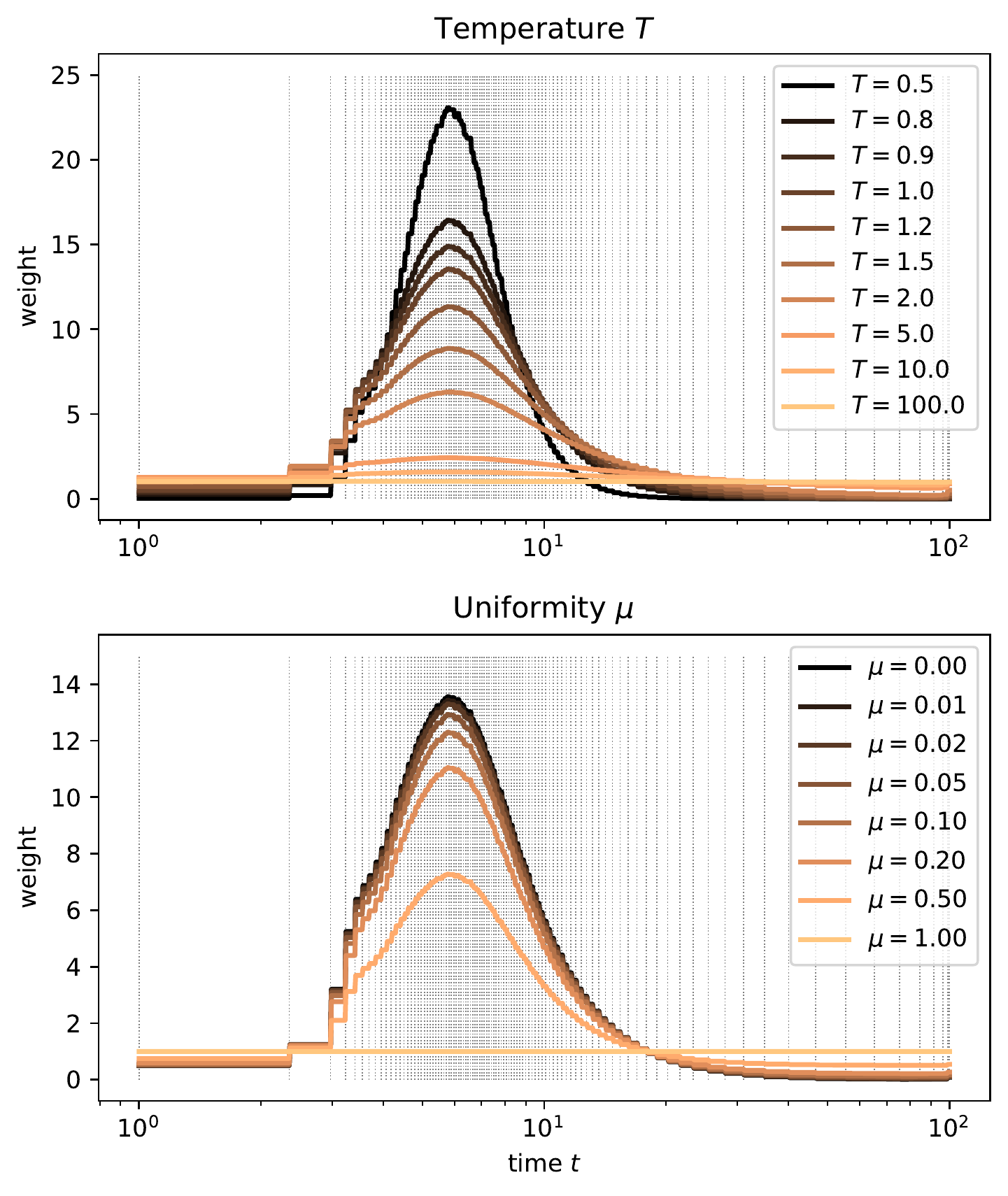}}
\caption{Effect of tuning the temperature (top) and the uniformity (bottom) of the time warping function from Figure~\ref{fig:time-warping}.}
\label{fig:temperature-uniformity}
\end{center}
\end{figure}

The effect of these manipulations is visualised in Figure~\ref{fig:temperature-uniformity}.

\section{Negative results}
\label{apx:negative-results}

We informally discuss some failed attempts to improve our results. This information is provided to help understand some of our design choices, and to aid researchers and practitioners who are interested in using these methods or investigating ways to improve them. Bearing in mind that context and details significantly impact experimental results, especially in machine learning research,  we expressly do not wish to discourage anyone from pursuing the ideas described here.

\subsection{Constraining the embedding parameters}
\label{apx:constraining-embedding-parameters}

Instead of normalisation, the embedding parameters can also be prevented from growing uncontrollably by adding a regularisation loss term. We experimented with L2 regularisation, margin penalties~\citep{donahue2019piano,dieleman2021variable}, as well as automatic adaptation of the L2 penalty weight to target unit variance for the embedding vector components~\citep{rezende2018generalized,dieleman2021variable}, but found that normalisation works best.

\subsection{Removing the time dependency}
\label{apx:removing-time-dependency}

In an effort to simplify the model architecture and make it resemble BERT even more strongly, we investigated models without time embedding (i.e. we removed the time embedding MLP in Figure~\ref{fig:architecture-diagram}). Since the model is solving a classification task, our hypothesis was that it could simply infer the noise level from the noisy input vectors, instead of requiring explicit knowledge of the timestep $t$. Unfortunately this significantly hurt performance. We suspect that this is a result of the relative scale of the embeddings and the noise (which varies greatly across noise levels), as well as the rescaling we apply to the noisy embeddings to ensure unit variance (see \S\ref{sec:architecture-and-hyperparameters}).

\subsection{Simplex diffusion}
\label{apx:simplex-diffusion}

We started out by lifting discrete token sequences into the space of categorical distributions over tokens, and performing diffusion in that space. Categorical distributions are nonnegative real-valued vectors whose components sum to 1, so they live on the simplex. We explored a tractable formulation of diffusion on the simplex, based on the Cox-Ingersoll-Ross process~\citep{cox1985theory}, as described by \citet{richemond2022categorical}.

We found that language modelling with this process is impeded by the uneven nature of the corruption process, which is a consequence of the high dimensionality of the simplex (corresponding to the number of tokens in the vocabulary $V$). In practice, the noise distribution is heavy-tailed and introduces frequent outliers, which make the corrupted vectors look like they correspond to the wrong tokens, even at very low noise levels. We were not able to circumvent this issue even with powerful Transformer models. We hypothesise that mitigating this issue requires modifying the corruption process so that it does not operate independently on all components. Formulating such a correlated process with a tractable transition density is non-trivial however, and this would complicate the model to a certain degree.

Both score interpolation (\S\ref{sec:interpolation}) and time warping (\S\ref{sec:time-warping}) were originally developed in the context of simplex diffusion, but we found these ideas to be more effective in combination with Gaussian diffusion in a Euclidean embedding space.

\end{document}